\documentclass[10pt,twocolumn,letterpaper]{article}

\usepackage[pagenumbers]{cvpr} 
\usepackage[pagebackref,breaklinks,colorlinks]{hyperref}

\usepackage[utf8]{inputenc}
\usepackage{amsmath}
\usepackage{amssymb}
\usepackage{graphicx}
\usepackage{tabularx}
\def\*#1{\mathbf{#1}}

\usepackage{caption}
\usepackage{subcaption}
\usepackage{multirow}
\usepackage{enumitem}
\usepackage{gensymb}
\usepackage{microtype}

\setitemize{noitemsep,topsep=0.5pt,parsep=0.5pt,partopsep=0.5pt,leftmargin=*}

\usepackage[capitalize]{cleveref}
\crefname{section}{Sec.}{Secs.}
\Crefname{section}{Section}{Sections}
\Crefname{table}{Table}{Tables}
\crefname{table}{Tab.}{Tabs.}

\def\cvprPaperID{2023} 
\def\confName{CVPR}
\def\confYear{2022}

\newcommand{\figref}[1]{\figurename~\ref{#1}}

\begin{document}

\title{Improving neural implicit surfaces geometry with patch warping}

\author{Fran\c{c}ois Darmon\textsuperscript{12}\quad Bénédicte Bascle\textsuperscript{1}\quad Jean-Clément Devaux\textsuperscript{1}\quad Pascal Monasse\textsuperscript{2}\quad Mathieu Aubry\textsuperscript{2} \\
\small \textsuperscript{1}Thales LAS France \quad
\small \textsuperscript{2}LIGM (UMR 8049), \'Ecole des Ponts, Univ. Gustave Eiffel, CNRS, Marne-la-Vall\'ee, France \\
\small \url{http://imagine.enpc.fr/\~darmonf/NeuralWarp/}
}
\maketitle

\begin{abstract}
    Neural implicit surfaces have become an important technique for multi-view 3D reconstruction but their accuracy remains limited. In this paper, we argue that this comes from the difficulty to learn and render high frequency textures with neural networks.
    We thus propose to add to the standard neural rendering optimization a direct photo-consistency term across the different views. Intuitively, we optimize the implicit geometry so that it warps views on each other in a consistent way. We demonstrate that two elements are key to the success of such an approach: (i)~warping entire patches, using the predicted occupancy and normals of the 3D points along each ray, and measuring their similarity with a robust structural similarity (SSIM); (ii)~handling visibility and occlusion in such a way that incorrect warps are not given too much importance while encouraging a reconstruction as complete as possible.
    We evaluate our approach, dubbed NeuralWarp, on the standard DTU and EPFL benchmarks and show it outperforms state of the art unsupervised implicit surfaces reconstructions by over 20\% on both datasets. Our code is available at \normalfont{\url{https://github.com/fdarmon/NeuralWarp}}
\end{abstract}

\section{Introduction}


Multi-view 3D reconstruction is the task of recovering the geometry of objects by looking at their projected views. Multi-View Stereo (MVS) methods rely on the photo-consistency of multiple views and typically provide the best results~\cite{colmap,vismvsnet}. However, they require a cumbersome multi-step procedure, first estimating then merging depth maps. Recent 3D optimization methods~\cite{dvr,nerf,idr,unisurf,volsdf,neus} avoid this issue by representing the surface implicitly and jointly optimizing neural networks encoding occupancy and color for all images, but their accuracy remains limited. In this work, we bridge these two types of approaches by optimizing multi-view photo-consistency for a geometry represented by implicit functions. We show that this enables our method to leverage high-frequency textures present in the input images that existing implicit methods struggle to represent, resulting in significant accuracy gains.

\begin{figure}[t]
    \centering
    \begin{subfigure}[t]{\columnwidth}
        \centering
        \includegraphics[page=1, width=0.65\linewidth]{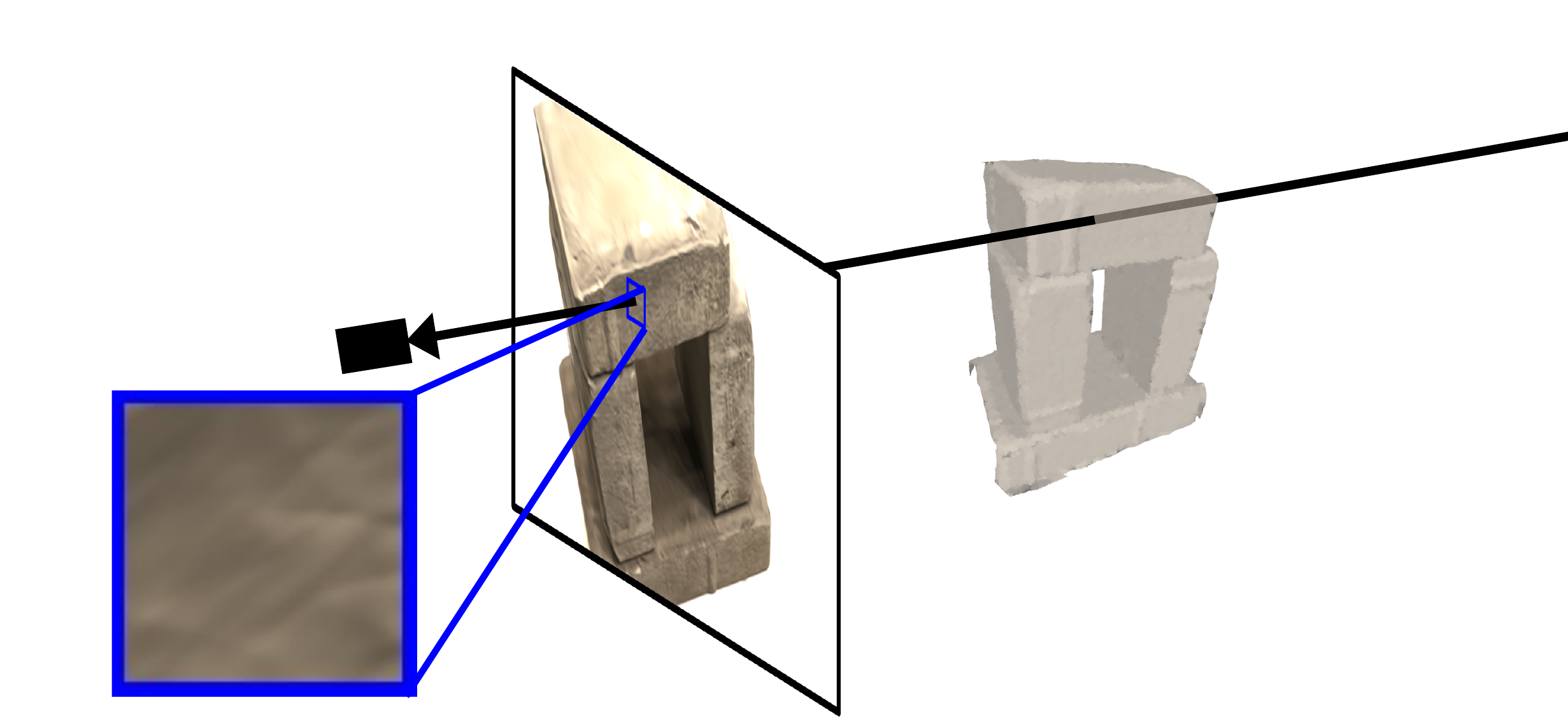}
        \includegraphics[width=0.34 \linewidth]{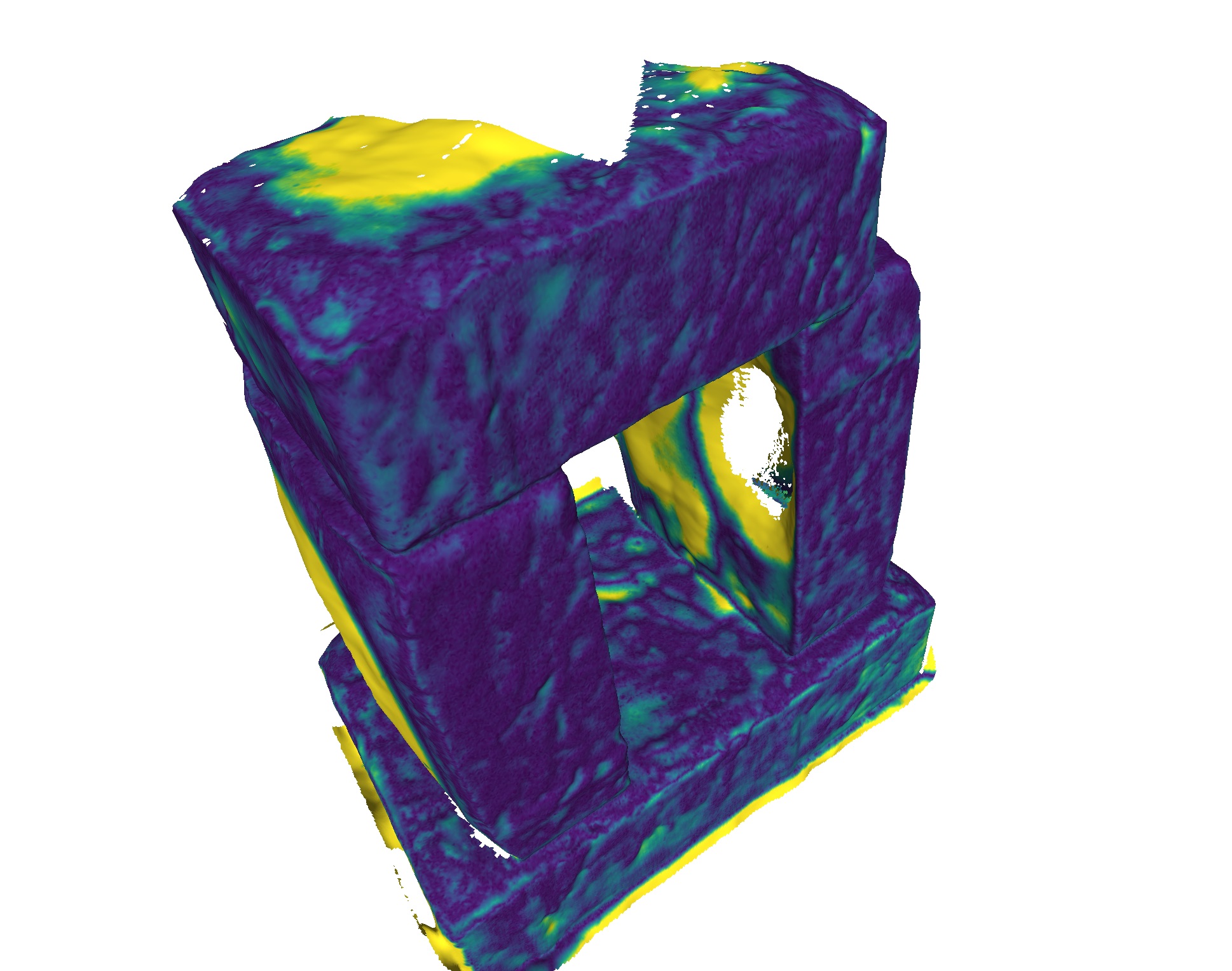}
        \caption{Volumetric rendering (left) and geometric error map (right).}
        \label{fig:teaser_a}
    \end{subfigure}
    \begin{subfigure}[t]{\columnwidth}
        \centering
        \includegraphics[page=2, width=0.65\linewidth]{figures/teaser_v2.pdf}
        \includegraphics[width=0.34\linewidth]{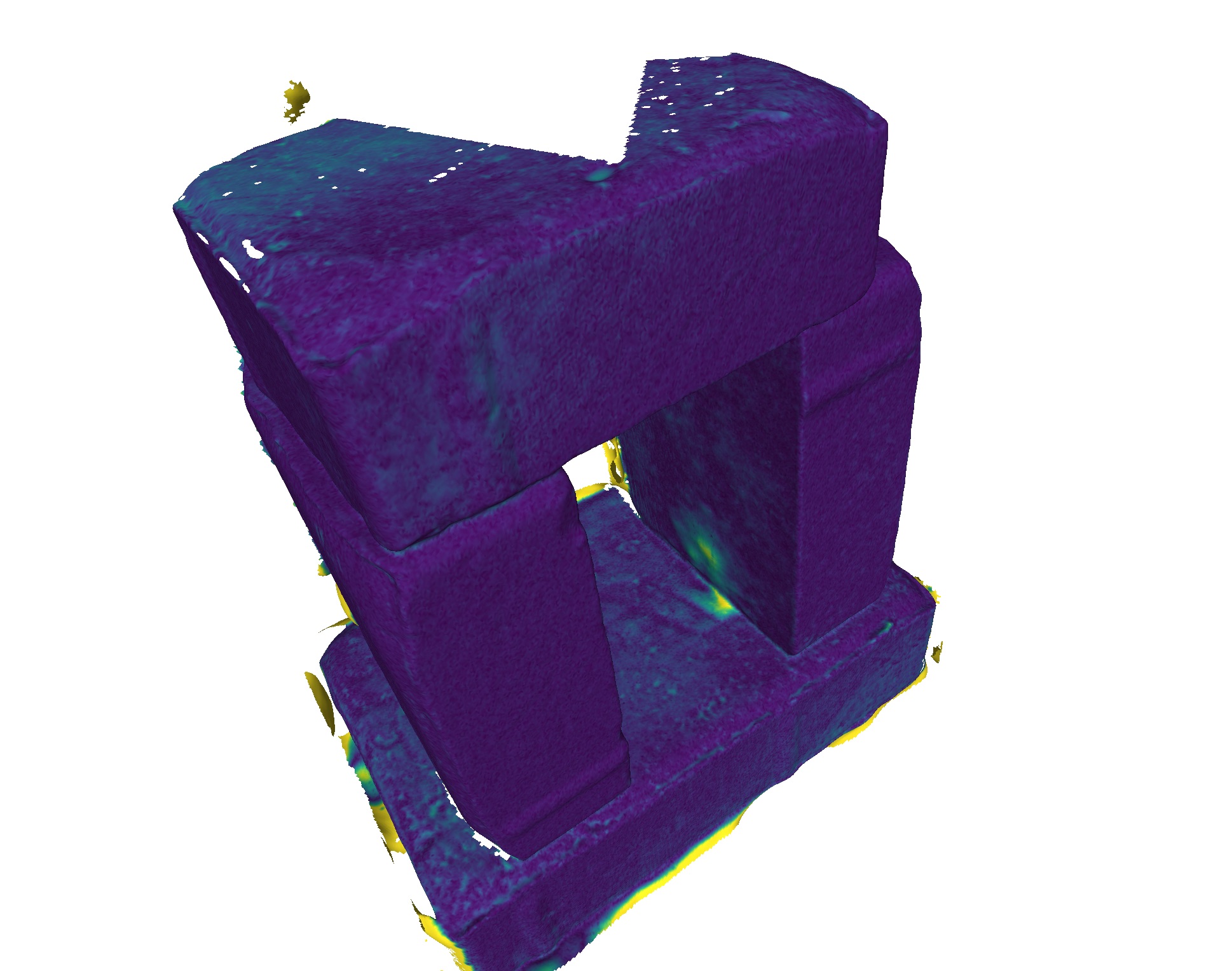}
        \caption{Our method's image warping (left) and geometric error map (right)}
        \label{fig:teaser_b}
    \end{subfigure}
    \caption{Standard neural implicit surface approaches jointly optimize a geometry and color network, but struggle to represent high frequency textures and therefore lack accuracy (top). We propose to additionally warp image patches with the implicit geometry, which allows to directly optimize photo-consistency between images and significantly improves the reconstruction accuracy (bottom).}
    \label{fig:teaser}
\end{figure}

The idea behind our approach is visualized on \figref{fig:teaser}. The top row shows a rendering and the geometric error map~(\ref{fig:teaser_a}) for a state of the art implicit method~\cite{volsdf}. The rendering fails at producing high frequency textures, resulting in low 3D accuracy. To overcome this limitation we use the original images, reprojecting them using the geometry described by the implicit occupancy function. This is shown on the bottom row~(\ref{fig:teaser_b}) where our warped patch includes high frequency texture. Consequently, we can optimize the geometry much more accurately, resulting in smaller geometric errors in the reconstruction.

Optimizing the implicit geometry for photo-consistency poses two main challenges. First, since we do not have perfectly Lambertian materials, directly minimizing the difference between colors is not meaningful and would lead to artefacts. We thus compare entire patches using a robust similarity (SSIM~\cite{ssim}) which requires performing patch warping using the implicit geometry. Building on the volumetric neural implicit surface framework, we start by sampling 3D points on the ray associated to each pixel in a reference image. We then propose to warp for each sampled point a source image patch to the reference image using a planar scene approximation, and finally combine all warped patches. Second, opposite to standard neural rendering methods that can associate a color to each 3D point, a warping-based approach must deal with the fact that many 3D points do not project correctly in the source view, e.g. are not visible or are occluded; this will typically happen for points sampled on any ray. We thus define for each reference image pixel and each source image a soft visibility mask. We then completely remove from the loss the contribution of pixels in the reference image that have no valid reprojection in any of the source views and, for the other pixels, weight the loss associated to each source view depending on how reliable the associated projection is. This downweights invalid reprojections, while encouraging a reconstruction as complete as possible.

We evaluate our method on the DTU~\cite{dtu} and EPFL~\cite{strecha} benchmarks. Our method outperforms current state-of-the-art unsupervised neural implicit surfaces methods by a large margin: the 3D reconstruction metrics are on average improved by $20\%$. We also show qualitatively that our image warps are able to capture high frequency details.\\ 

\noindent To summarize, we present:
\begin{itemize}
\item a method to warp patches using implicit geometry;
\item a loss function able to handle incorrect reprojections;
\item an experimental evaluation demonstrating the very significant accuracy gain on two standard benchmarks and validating each element of our approach.
\end{itemize}

\section{Related Work}

Multi-view 3D reconstruction, the task of recovering the 3D geometry of a scene from 2D images, is a long standing problem in computer vision. We focus on the calibrated scenario where both camera calibrations are known. In this section, we first review Multi-View Stereo (MVS) methods, then 
neural implicit methods that optimize neural networks for image rendering and finally 
methods that use projections in multiple views with implicit methods. 

\vspace{-8pt}
\paragraph{Multi-view stereo (MVS):}
Classical MVS approaches use 3D representations such as 3D point clouds like PMVS~\cite{pmvs}, voxel grids~\cite{seitz1999photorealistic,kutulakos2000theory} or depth maps~\cite{Zheng_2014_CVPR,gipuma,colmap}. A more detailed overview can be found in~\cite{furukawa2015multi}. Depth map based methods are arguably the most common, 
with the widespread usage of COLMAP~\cite{colmap}. This approach relies on a graphical model and optimizes depth and normal maps in a multi-step optimization. It ends with a depth map fusion step that outputs a point cloud, which can be further processed with a meshing algorithm~\cite{spsr,meshing_enpc}. 
 Deep learning has also been successfully applied to MVS estimation. Most methods output depth map estimates~\cite{mvsnet,rmvsnet,pointmvsnet,cvpmvsnet,vismvsnet}, but some also produce voxels~\cite{surfacenet,atlas} or point clouds~\cite{deltas}. These methods achieve impressive results on multiple benchmarks~\cite{dtu,tanks} but they are supervised and trained on specific datasets~\cite{dtu,blendedmvs}. Unsupervised deep MVS methods have also been introduced~\cite{mvs2,jdacs,perso3dv} but their performances are still limited compared to supervised versions. Our method is fully unsupervised and requires neither training data nor pretrained networks, but has high performance. 

\vspace{-8pt}
\paragraph{Neural implicit surfaces:}
Recently, new implicit representations of 3D surfaces with neural network were introduced. 
The surface is represented by a neural network which will output either 
 an occupancy field~\cite{occnet,convoccnet} or a Signed Distance Function (SDF)~\cite{deepsdf}.  
These representations are used to perform multi-view reconstruction following two different paradigms~\cite{unisurf}: surfacic~\cite{dvr,idr,mvsdf} and volumetric~\cite{nerf,unisurf,volsdf,neus}. Surfacic approaches compute the surface then backpropagate through this step with implicit differentiation~\cite{neurlevelsets}. They are hard to optimize and typically require additional supervision: silhouette masks in~\cite{dvr, idr} or the output of a pretrained depth map estimator~\cite{vismvsnet} in MVSDF~\cite{mvsdf}. Volumetric approaches were introduced in NeRF~\cite{nerf}. The latter combines classical volumetric rendering~\cite{kajiya1984ray} with implicit functions to produce high quality renderings of images. The main focus of NeRF~\cite{nerf} was the quality of rendering, therefore the geometry was not evaluated. Further work adapted the geometric output~\cite{unisurf,volsdf,neus} to make it better suited for surface extraction. UNISURF~\cite{unisurf} uses an occupancy network~\cite{occnet} whereas VolSDF~\cite{volsdf} and NeuS~\cite{neus} use an SDF~\cite{deepsdf}. We build our method on VolSDF~\cite{volsdf} but we believe it could be adapted to fit any volumetric neural implicit framework. 

\begin{figure*}[ht]
    \includegraphics[page=1,width=\linewidth]{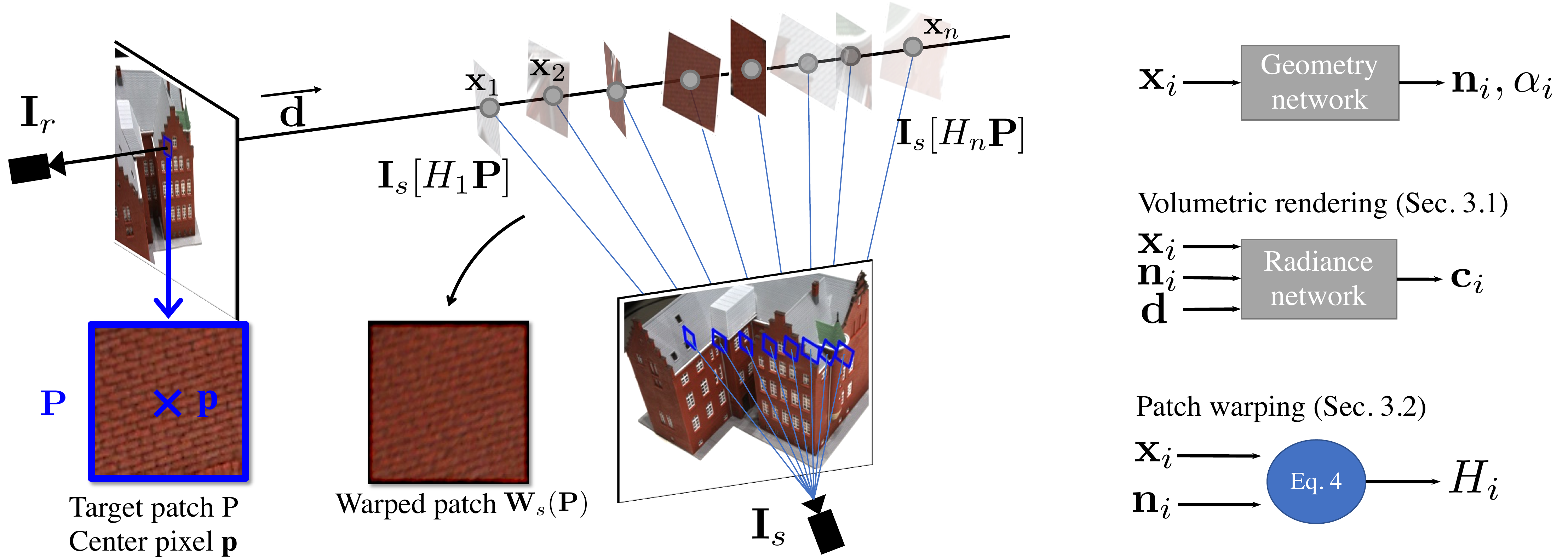}
    \caption{Approach overview. We combine volumetric rendering with a new patch warping technique. Both approaches aggregate color from points sampled along the camera ray: radiance predicted by the radiance network for volumetric rendering and patch extracted from source views for our patch warping.}
    \label{fig:method}
\end{figure*}

\vspace{-8pt}
\paragraph{Image warping and neural implicit surfaces:}

In this work we combine the color matching idea of traditional MVS with neural implicit surfaces. Closest to this idea is MVSDF~\cite{mvsdf} that also uses a loss based on correspondences and works on accurate geometry optimization. However, it optimizes consistency between CNN features and the optimization requires a network pretrained on multi-view datasets~\cite{blendedmvs}. Our approach does not require any pretrained network and we show that it outperforms MVSDF. 

The idea of projecting information from source views to 3D then using the neural radiance field framework to render a target view has also been used in learning-based approaches~\cite{pixelnerf,ibrnet,srf, grf, mvsnerf, metanlr}. These approaches train on multiple scenes networks that take as input features from the source views aggregated at a given 3D point and output radiance and occupancy for this point. They focus however on the generalization to new scenes and the quality of rendered views, but the quality of their predicted geometry has not been evaluated, to the best of our knowledge. On the contrary, we focus on the optimization framework, i.e., we do not train our network on several scenes, and we optimize the quality of geometry. 

\section{Method}
In this section, we present our technical contributions.
Section~\ref{sec:vol} introduces the volumetric rendering framework on which we build. Section~\ref{sec:warp} explains how we warp a patch from a source image to a target image given a 3D scene represented by a geometry network predicting occupancy for each 3D point. Section~\ref{sec:masks} discusses questions related to visibility and how we mask invalid points during the optimization. Finally, Section~\ref{sec:impl} presents our full optimization. An overview of our approach and notations can be seen in Figure~\ref{fig:method}.

\subsection{Volumetric rendering of radiance field}
\label{sec:vol}

Neural volumetric rendering was introduced in~\cite{nerf} for novel view rendering. The idea is to represent the characteristics of a 3D scene with two implicit functions that are approximated with neural networks. The geometry network encodes the geometry of the scene, we use Signed Distance Field (SDF) encoding~\cite{volsdf, neus}. The radiance network encodes the color emitted by any region in space in all directions. The idea is to optimize the two neural networks together so that rendering the associated scene reconstructs a set of given views of the scene.  
Let us consider a reference image $\*I_r$. The two networks are optimized using $\ell1$ loss between the colors in reference image $\*{I}_r[\*{p}]$ and the volumetric rendering $R[\*p]$ for a pixel $\*p$:
\begin{equation}
    \mathcal{L}_{vol} = \sum_{\*p} |\*{I}_r[\*{p}] - \*R[\*p]|.
    \label{eq:loss_vol}
\end{equation}
The rendered color $\*R[\*p]$ is computed from both networks in a differentiable way with respect to their parameters using volumetric rendering~\cite{kajiya1984ray}. 
Let $\*x_i, i=1\dots n$ be an ordered set of points sampled along the ray going through the reference camera center and the pixel $\*p$.\footnote{For simplicity, we drop from the point notations the dependency on the pixel to render $\*p$ and the reference image index $r$.} The rendering of the scene at pixel $\*p$ is approximated as a weighted sum of the radiance $\*c_i$ at each point using weights computed from the geometry network. 
Intuitively, the color $\*c_i$ will contribute to the rendering if $\*x_i$ has a high density and if no point on the ray between $\*x_i$ and the reference camera has a high density value. Formally, $\*c_i = \*c(\*x_i, \*n_i, \*d)$ is the radiance computed with the radiance network $\*c$ in ray direction $\*d$, at the points $\*x_i$ of surface normals $\*n_i$, computed by differentiation of the geometry network at the different positions $\*x_i$. The rendered color is approximated with an alpha blending of the $\*c_i$.
\begin{align}
    \*R[\*p] &=\sum_{i=1}^N \alpha_i\prod_{j < i}(1 - \alpha_j)\*c_i,
    \label{eq:vol_render}
\end{align}
where we consider for simplicity that the geometry network $\alpha$ outputs occupancy values $\alpha_i = \alpha(\*x_i)$ between $0$ and $1$. In practice, our geometry network outputs an SDF, and we refer to~\cite{volsdf} for a detailed explanation of the mapping of SDF to occupancy values. Eq. \eqref{eq:vol_render} is the discrete approximation of an integral along the camera ray. Therefore, the choice of sampling points $\*x_i$ is a key element, discussed in~\cite{nerf,unisurf,volsdf, neus}. Those methods improve the geometry estimation by focusing on the sampling, but the reconstructions are still worse than the traditional MVS techniques. Our hypothesis is that it comes from the difficulty of the radiance network to represent high frequency textures (see \figref{fig:teaser}). 

\subsection{Warping images with implicit geometry}
\label{sec:warp}

Instead of memorizing all the color information present in the scene with the radiance network, we propose to directly warp images onto each other relying only on the geometry network. We consider a reference image $\* I_r$ and a source image $\* I_s$. Similar to the above section, we want to obtain the color of a pixel $\*p$ or a patch centered around $\*p$ using the occupancies $\alpha_i, i=1\dots N$ from the geometry network but this time using colors from projections from the source image and not the one predicted by the radiance network. In this section, we assume these projections and their colors are well defined, and deal with the general case in Section~\ref{sec:masks}. We start by explaining how a source image can be warped to a target image pixel-by-pixel, which we refer to as pixel warping. We then extend this idea to warping full patches, a classical idea in MVS~\cite{pmvs,Zheng_2014_CVPR,gipuma,colmap}.

\paragraph{Pixel warping:} 
Instead of using a radiance network to compute the color of each 3D point $\*x_i$ on the ray associated to pixel $\*p$, we use the color of their projection in source image. Formally, we define the warped value of $\*p$ from source image $s$ as:
\begin{equation}
    \*W_s[\*p] = \sum_{i=1}^N \alpha_i\prod_{j < i}(1 - \alpha_j)\*I_s[\pi_s(\*x_i)],
    \label{eq:pix_warp}
\end{equation}
where $\*I_s[\pi_s(\*x)]$ denotes the bilinear interpolation of colors from $\*I_s$ at the point $\pi_s(\*x)$ where the 3D point $\*x$ projects in $\*I_s$. In this section, we assume that every 3D points has a valid projection in source image so that $\*I_s[\pi_s(\*x)]$ is always defined. 
Eq.~\eqref{eq:pix_warp} is similar to Eq.~\eqref{eq:vol_render} but the color comes from pixel values in source images instead of network predictions. Intuitively, the warped value is a weighted average of the source image colors along the epipolar line. Similar to Eq.~\ref{eq:loss_vol}, one could optimize the geometry using a $\ell1$ loss function $\ell = \sum_\*p |\*W_s[\*p] - I_r[\*p]|$. However, 
this does not model changes in intensity related to the camera viewpoint, and in particular specularities, 
and can create artifacts in the reconstruction. One solution to this issue is to use a robust patch-based photometric loss function. 

\vspace{-8pt}
\paragraph{Patch warping:} 

We now explain how to warp entire patches instead of single pixels, by locally approximating the scene at each point $\*x_i$ as a plane in a way similar to standard techniques used in classical MVS~\cite{pmvs}. 
Let $\*n_i$ be the surface normal at $\*x_i$, which can be computed with automatic differentiation of the geometry network at $\*x_i$. Let $H_i$ be the homography between $\*I_r$ and $\*I_s$ induced by the plane through $\*x_i$ of normal $\*n_i$. It can be computed as a $3\times3$ matrix acting on 2D homogeneous coordinates:
\begin{equation}
    H_i = K_s\left(R_{rs} + \frac{\*t_{rs}\*n_i^TR_r^T}{\*n_i^T(\*x_i + R_r^T\*t_r)}\right)K_r^{-1},
    \label{eq:homs}
\end{equation}
where $K_r$ and $K_s$ are the internal calibration matrices of reference and source cameras, $(R_{rs}, \*t_{rs})$ is the relative motion from $\*I_r$ to $\*I_s$ represented by a $3\times3$ rotation matrix $R_{rs}$ and a 3D translation vector $\*t_{rs}$ and $(R_r, \*t_r)$ is the pose of reference frame in world coordinates with the same representation. This homography associates any pixel $\*q$ in $\*I_r$ to a pixel $H\_i*q$ in $\*I_s$. 

With a slight abuse we extend this notation to patches: for a patch $\*P$ centered around pixel~$\*p$,  we write $H_i\*P$ the application of the homography to all pixels of the patch and $\*I_s[H_i\*P]$ the color interpolated at those locations in $\*I_s$. Intuitively, $H_i\*P$ is the location of a patch in source image that would correspond to $\*P$ in reference frame if the true geometry was a plane of normal $\*n_i$ passing through $\*x_i$. We can now average the patches corresponding to each $\*x_i$ in a manner similar to Eq.~$\eqref{eq:vol_render}$ to produce a warped patch $\*W_{s}[\*P]$:
\begin{equation}
    \*W_{s}[\*P] = \sum_{i=1}^N \alpha_i\prod_{j < i}(1 - \alpha_j)\*I_s[H_i \*P].
    \label{eq:patch_warp}
\end{equation}
In all our experiments we follow COLMAP~\cite{colmap} and use a patch size of $11\times11$. Note that using a patch size of $1$ would provide the same equation as~Eq.~\eqref{eq:pix_warp}. 

\begin{table*}[ht]
    \centering
    \small
    \begin{tabularx}{\linewidth}{c|X X X X X X X X X X X X X X X | c}
    Scan & 24 & 37 & 40 & 55 & 63 & 65 & 69 & 83 & 97 & 105 & 106 & 110 & 114 & 118 & 122 & \bf Mean \\
    \hline
    IDR~\cite{idr} & 1.63 & 1.87 & 0.63 & 0.48 & 1.04 & 0.79 & 0.77 & 1.33 & 1.16 & 0.76 & 0.67 & 0.90 & 0.42 & 0.51 & 0.53 & 0.90 \\
    MVSDF~\cite{mvsdf}* & 0.83 & 1.76 & 0.88 & 0.44 & 1.11 & 0.90 & 0.75 & 1.26 & 1.02 & 1.35 & 0.87 & 0.84 & 0.34 & 0.47 & 0.46 & 0.88 \\
    COLMAP~\cite{colmap} & 0.45 & 0.91 & 0.37 & 0.37 & 0.90 & 1.00 & 0.54 & 1.22 & 1.08 & 0.64 & 0.48 & 0.59 & 0.32 & 0.45 & 0.43 & 0.65 \\
    \hline
    NeRF~\cite{nerf} & 1.90 & 1.60 & 1.85 & 0.58 & 2.28 & 1.27 & 1.47 & 1.67 & 2.05 & 1.07 & 0.88 & 2.53 & 1.06 & 1.15 & 0.96 & 1.49 \\
    UNISURF~\cite{unisurf} & 1.32 & 1.36 & 1.72 & 0.44 & 1.35 & \underline{0.79} & \underline{0.80} & 1.49 & 1.37 & 0.89 & \bf 0.59 & 1.47 & 0.46 & \underline{0.59} & 0.62 & 1.02 \\
    NeuS~\cite{neus} & 1.37 & \underline{1.21} & \underline{0.73} & \underline{0.40} & \underline{1.20} & \bf 0.70 & \bf 0.72 & \bf 1.01 & \underline{1.16} & 0.82 & \underline{0.66} & 1.69 & \bf 0.39 & \bf 0.49 & \bf 0.51 & 0.87 \\
    VolSDF~\cite{volsdf} & \underline{1.14} & 1.26 & 0.81 & 0.49 & 1.25 & \bf 0.70 & \bf 0.72 & 1.29 & 1.18 & \underline{0.70} & \underline{0.66} & \underline{1.08} & 0.42 & 0.61 & \underline{0.55} & \underline{0.86} \\
    NeuralWarp (ours) & \bf 0.49 & \bf 0.71 & \bf 0.38 & \bf 0.38 & \bf 0.79 & 0.81 & 0.82 & \underline{1.20} & \bf 1.06 & \bf 0.68 & \underline{0.66} & \bf 0.74 & \underline{0.41} & 0.63 & \bf 0.51 & \bf 0.68 \\

    \end{tabularx}
    \caption{Quantitative comparison on DTU. All the results are the ones reported in the original papers, except NeRF, IDR and COLMAP results that come from~\cite{unisurf}. Note that MVSDF results (*) are not directly comparable since they use a custom filtering whereas results with all other methods have been cleaned using the visual hull. The bottom part of the table compares our result with neural implicit surfaces approaches that do not use additional inputs (masks, supervised depth estimation). \textbf{Bold} results have the best score and \underline{underlined} the second best. Our method outperforms existing work by a large margin, more than 20\% on the mean metrics.}
    \vspace{-8pt}
    \label{tab:dtu}
\end{table*}

\subsection{Optimizing geometry from warped patches}
\label{sec:masks}

We now want to define a loss based on the warped patches to optimize the geometry. This cannot be done by using directly~\eqref{eq:patch_warp} to warp patches and maximizing SSIM. Indeed, we assumed that all 3D points $\*x_i$ on the camera ray have a valid projection in the source image, but in practice this is not true for many points, which may project outside of the source image for example. In that case $\*I_s[H_i\*P]$ is not defined, and we instead use a constant (gray) padding color in~\eqref{eq:patch_warp}. 
This will of course affect the quality of the warped patch $\*W_s[\*P]$. 
Intuitively, if all non-valid 3D points on a ray are far from the implicit surface seen in the reference image, the padding value will contribute very little to the final warped patch which can be used in the loss. On the contrary, if there are invalid points near the implicit surface, the warped patch becomes invalid and should not be used in the loss. We formalize this intuition by assigning to a patch $\* P$ centered around pixel $\*p$  in the reference image a mask value $M_s[\*P]\in[0,1]$ for each source image $s$. In the rest of the section, we first explain how we define our loss for a given reference image based on the validity masks $M_s$ associated with each source image; We then explain how we define the validity masks.

\vspace{-8pt}
\paragraph{Warping-based loss:} 
We start by selecting for a given reference image the set $\mathcal{V}$ of patches we consider in the loss. Since we want to discard patches $\* P$ for which no source image gives ant reasonable warp, as quantified by $M_s(\*P)$, we define it as $\mathcal{V}=\{\*P : \sum_s M_s(\*P)> \epsilon\}$. In practice, we use $\epsilon=0.001$ in all of our experiments. 
We then define our warping-based loss such that every valid patch in the reference image is given the same weight, but also such that invalid warping are given less weight:
\begin{equation}
    \mathcal{L}_\text{warp} = \sum_{\* P\in \mathcal{V}} \dfrac{\sum_{s\in \mathcal{S}}M_s[\*P]\,d(\*I_r[\*P], \*W_s[\*P])}{\sum_{s\in \mathcal{S}}M_s[\*P ]}
    \label{eq:loss_warp}
\end{equation}
where $\*I_r[\*P]$ is the color patch $\*P$ in $\*I_r$ and $d$ is a photometric distance between image patches. We use for $d$ the SSIM~\cite{ssim}, except for our ablation where we use $\ell1$.

\vspace{-8pt}
\paragraph{Validity masks:}
We now explain how we define the validity mask $M_s$. We consider two reasons for warps not being valid, hence two masks: (i)~a projection mask $M^{\text{proj}}_s$ for cases in which the projection is not valid for geometric reasons, and (ii)~an occlusion mask $M^{\text{occ}}_s$ for cases where the patch is occluded by the reconstructed scene in the source image. The final mask is the product of both: $M_s[\*P]$ = $M^{\text{proj}}_s[\*P]M^{\text{occ}}_s[\*P]$.


To define the projection mask, we introduce a binary indicator $V_i^s$ which is $0$ when the projection associated with $\*x_i$ is not valid and $1$ otherwise. The projection can be invalid for three reasons: first, when the projection of the point in the source view is outside the source image; second, when the reference and source views are on two different sides of the plane defined by $\*x_i$ and the normal $\*n_i$; third, when a camera center is too close to the plane defined by $\*x_i$ and the normal $\*n_i$, for which we use a threshold of 0.001 in practice.  
We obtain $M^{\text{proj}}_s[\*P]$ by averaging the validity indicators for all 3D points sampled on the ray associated to $\*P $ weighted by the $\alpha$ values:
\begin{equation}
    M^{\text{proj}}_s[\*P] = \sum_{i=1}^N \alpha_i\prod_{j < i}(1 - \alpha_j)V_i^s
    \label{eq:reproj_mask}
\end{equation}
Note that~\eqref{eq:reproj_mask} produces a soft mask value between $0$ and $1$, which is necessary to make it differentiable with respect to the $\alpha$ factors. We found such a property to be important in practice. 

To define the occlusion mask $M^{\text{occ}}_s $ we check whether there are occupied regions on the ray between the points $\*x_i$ and the source camera center. 
We compute how occluded a 3D point $\*x$ is with its transmittance in source view: $T_s(\*x) = 1 -  \prod_{k=1}^{N}(1 - \alpha^s_{k})$, where the $\alpha^s_{k}$ are the occupancy values predicted by the geometry network on 3D points sampled on the ray from $\*x$ to the center of view $s$. Intuitively, $T_s(\*x)$ is close to $1$ if there is no point with a large density between the source image and $\*x$, otherwise it is close to $0$.
We could average the $T_s(\*x_i), i=1\dots N$ in the same manner as~Eq.~\eqref{eq:reproj_mask} but this would require computing the transmittance $T_s$ for every point on the ray. For computational efficiency we instead choose to compute an intersection point on the ray corresponding to the patch $\* P$ in the reference view and evaluate transmittance in the source views on this point only:
\begin{equation}
    M^{\text{occ}}_s[\*P] = T_s\left(\sum_{i=1}^N \alpha_i\prod_{j < i}(1 - \alpha_j)\*x_i\right)
\end{equation}
%
This mask is again soft because $T_s$ outputs a continuous value in range $[0,1]$, which helps handling thin surfaces occlusion: if a ray comes close to a surface without being strictly occluded, it will still have a lesser influence on the warping loss compared to a ray that is far from any surface. 

\begin{figure*}[ht]
  \begin{subfigure}{0.23\textwidth}
    \includegraphics[width=\textwidth, trim={0 50 0 0}, clip]{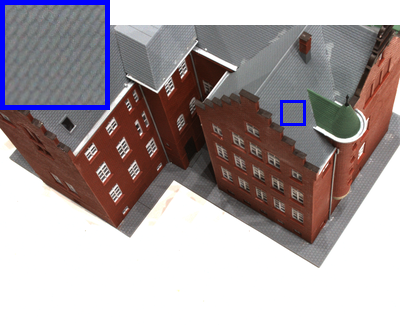}
  \end{subfigure}
  \begin{subfigure}{0.23\textwidth}
    \includegraphics[width=\textwidth, trim={0 50 0 0}, clip]{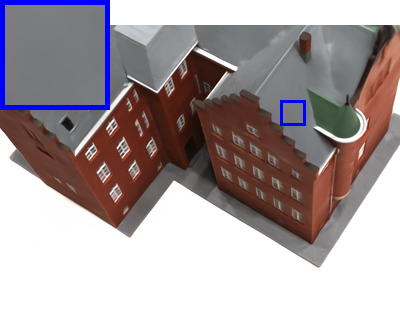}
  \end{subfigure}
  \begin{subfigure}{0.23\textwidth}
    \includegraphics[width=\textwidth, trim={0 50 0 0}, clip]{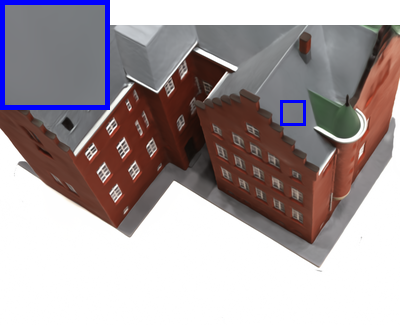}
  \end{subfigure}
  \begin{subfigure}{0.25\textwidth}
    \includegraphics[width=\textwidth, trim={0 50 0 0}, clip]{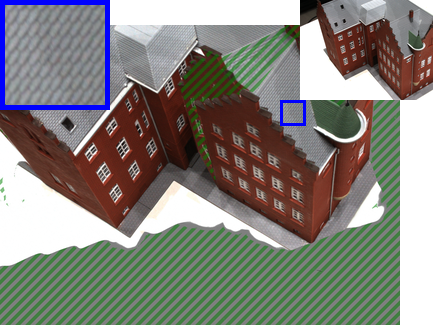}
  \end{subfigure}

  \begin{subfigure}{0.23\textwidth}
    \includegraphics[width=\textwidth, trim={0 50 0 0}, clip]{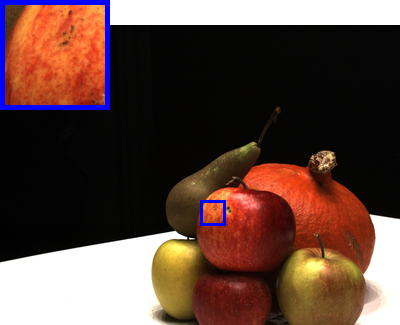}
    \caption*{Ground truth image}
  \end{subfigure}
  \begin{subfigure}{0.23\textwidth}
    \includegraphics[width=\textwidth, trim={0 50 0 0}, clip]{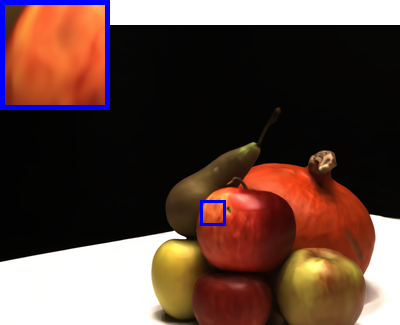}
    \caption*{VolSDF~\cite{volsdf} volume rendering}
  \end{subfigure}
  \begin{subfigure}{0.23\textwidth}
    \includegraphics[width=\textwidth, trim={0 50 0 0}, clip]{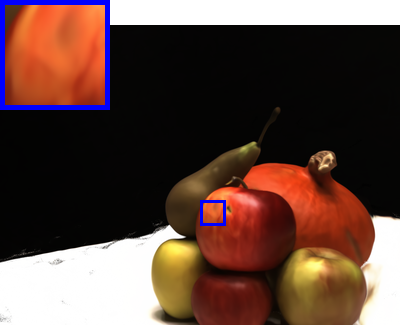}
    \caption*{\centering Our volume rendering using our radiance network}

  \end{subfigure}
  \begin{subfigure}{0.25\textwidth}
    \includegraphics[width=\textwidth, trim={0 50 0 0}, clip]{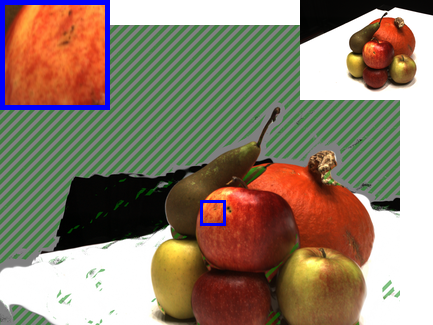}
    \caption*{\centering Ours warping from a source image and validity mask in green}
  \end{subfigure}
  \caption{Examples of DTU renderings using VolSDF~\cite{volsdf} and our method, both using our implementation. For both methods, volumetric rendering cannot produce high frequency texture whereas image warping can precisely recover them. (better viewed in electronic version)}
  \vspace{-8pt}
  \label{fig:dtu}
\end{figure*}

\subsection{Optimization details}
\label{sec:impl}

We now explain the details of our method. We first present out full loss and optimization. We then detail the network architecture. We finally discuss how we selected the source images for each reference image. 

\vspace{-8pt}
\paragraph{Full optimization:}

We optimize the geometry and radiance networks to minimize the sum of the volumetric rendering loss (Eq.~\ref{eq:loss_vol}) and the patch warping loss (Eq.~\ref{eq:loss_warp}). In order to encourage the geometry network to output a function similar to a signed distance field, we also add the eikonal loss $\mathcal{L}_\text{eik}$~\cite{eikonal} which is minimum when the gradient of the output function at each point in space is of norm $1$. This results in the following complete loss:
\begin{equation}
    \mathcal{L} = \lambda_\text{vol}\mathcal{L}_\text{vol} + \lambda_\text{warp}\mathcal{L}_\text{warp} + \lambda_\text{eik}\mathcal{L}_\text{eik}
\end{equation}
where $\lambda_\text{vol}$, $\lambda_\text{warp}$ and $\lambda_\text{eik}$ are scalar hyperparameters.
We use $\lambda_\text{vol} = 1$ and $\lambda_\text{eik} = 0.1$ for all experiments. We first optimize the networks  using $\lambda_\text{warp} = 0$ in the same setting as VolSDF. After $100k$ iterations with a learning rate exponentially decayed from $5\mathrm{e}{-4}$ to $5\mathrm{e}{-5}$, we finetune for another $50k$ iterations using $\lambda_\text{warp} = 1$ and a fixed learning rate of $1\mathrm{e}{-5}$. The networks are initialized with the sphere initialization of~\cite{sal}. 
We start optimizing with volumetric rendering only because the normals are initially too noisy to compute meaningful homographies. During the first phase of training, without patch warping, we train with batches of $1024$ pixels, but we finetune patch warping on batches of $512$ patches due to GPU memory constraints. Also, we do not backpropagate the loss through the homography parameters in equation \eqref{eq:homs} since we noticed it leads to unstable optimization but the geometry network is still optimized through the $\alpha$ factors of Eq.~\eqref{eq:patch_warp}.   

\vspace{-8pt}
\paragraph{Architecture:}
We use the same architecture as concurrent works~\cite{unisurf, volsdf}. Both radiance and geometry networks are Multi-Layer Perceptrons (MLP). The geometry network has 8 layers with 256 hidden units. The radiance network has 4 layers with 256 hidden units. Similar to~\cite{idr,unisurf,volsdf} we encode 3D position using positional encoding with 6 frequencies and viewing direction with 4 frequencies.

\vspace{-8pt}
\paragraph{Rays sampling:} We follow VolSDF~\cite{volsdf} in choosing the points $\*x_1\dots \*x_N$ on the camera rays with a small modification. We first estimate the opacity function with the algorithm introduced by VolSDF, we then sample $N = 64$ points on the camera ray with $90\%$ sampled from the opacity distribution as in VolSDF but the remaining $10\%$ sampled uniformly along the whole camera ray. 

\vspace{-8pt}
\paragraph{Choice of source images:}

Our method uses a set of $19$ source images $\mathcal{S}$ following COLMAP~\cite{colmap} for each reference image. Those source images must be carefully chosen since very similar viewpoints will carry little geometric information and very different viewpoints will have few common points. We first build a sparse point cloud with a Structure from Motion software~\cite{colmap_sfm}, we then compute for each image pair the total number of sparse points observed by both (co-visible points) and remove the pairs for which more than 75\% of the co-visible points are observed with a triangulation angle below $5\degree$ . We finally select the $19$ top views in number of co-visible points. 

\begin{figure}
    \centering
    \includegraphics[width=0.45\linewidth, trim={0, 75, 0, 0}, clip ]{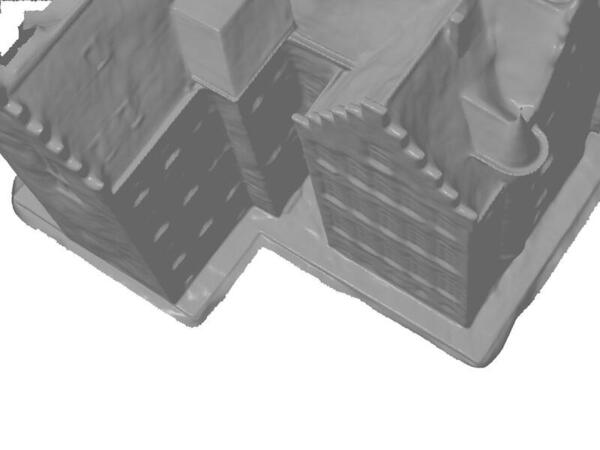}
    \includegraphics[width=0.45 \linewidth, trim={0, 75, 0, 0}, clip ]{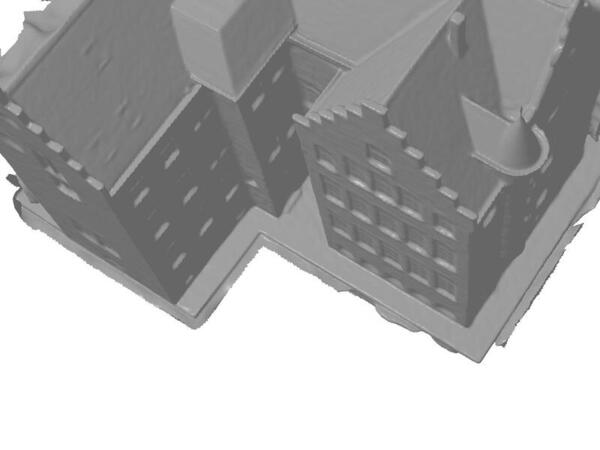}
    \includegraphics[width=0.45 \linewidth, trim={0, 75, 0, 0}, clip ]{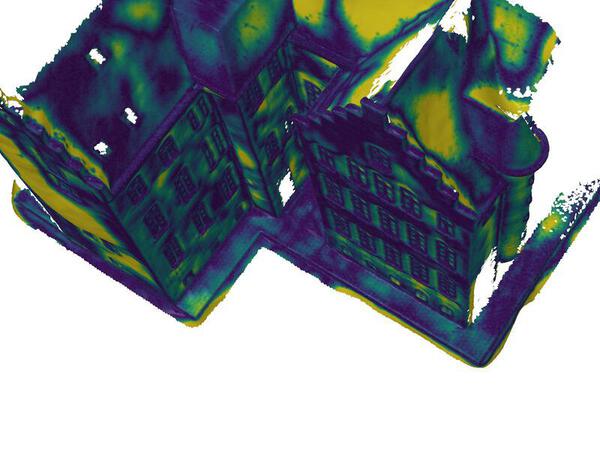}
    \includegraphics[width=0.45 \linewidth, trim={0, 75, 0, 0}, clip ]{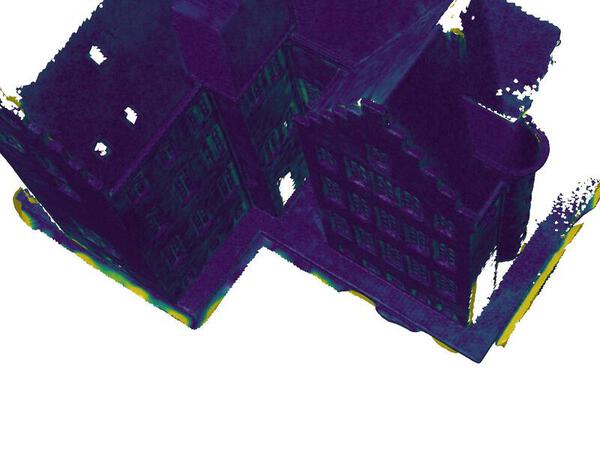}
    \includegraphics[width=0.45 \linewidth,trim={200, 0, 0, 170}, clip]{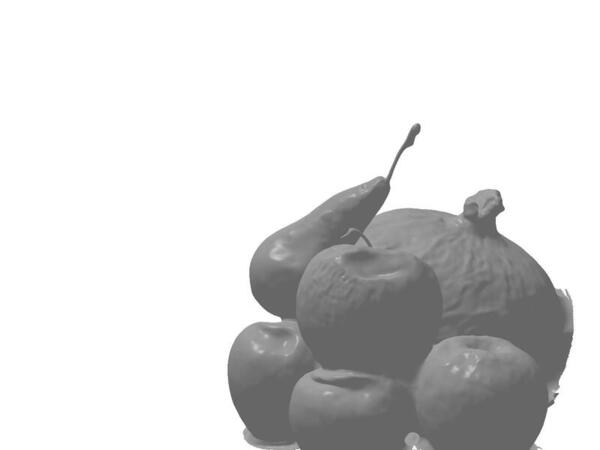}
    \includegraphics[width=0.45 \linewidth,trim={200, 0, 0, 170}, clip]{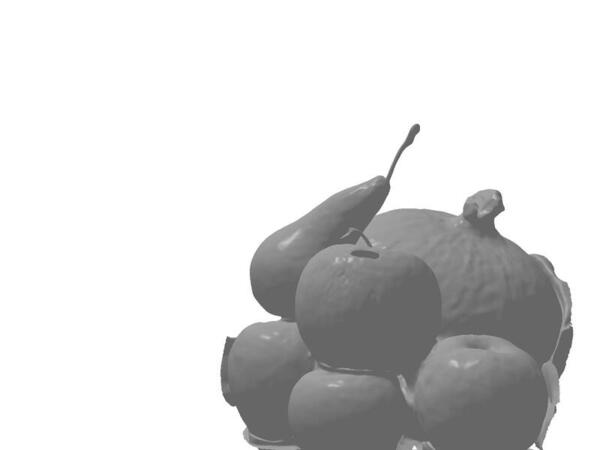}
    \includegraphics[width=0.45 \linewidth,trim={200, 0, 0, 170}, clip]{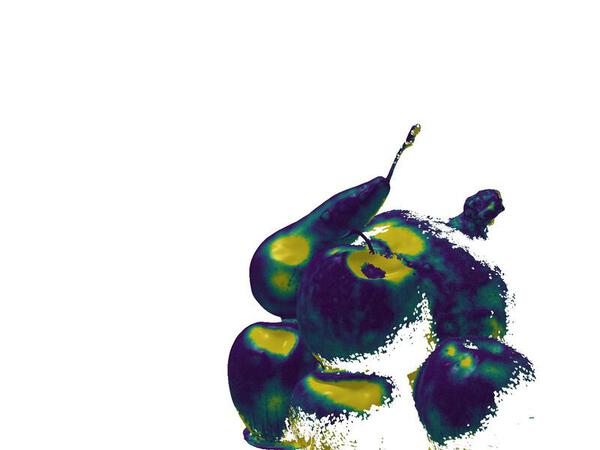}
    \includegraphics[width=0.45 \linewidth,trim={200, 0, 0, 170}, clip]{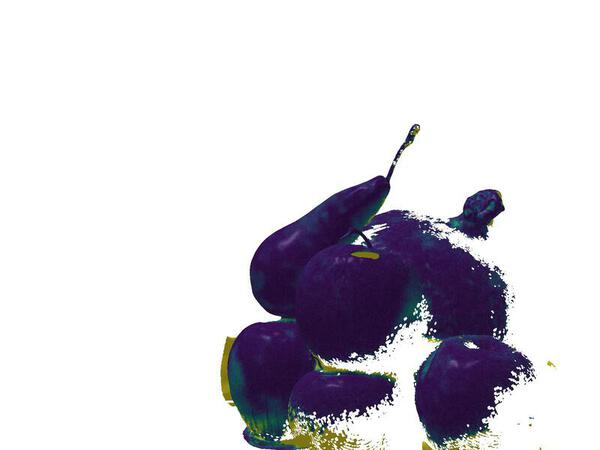}
    \begin{subfigure}{0.45\linewidth}
        \caption*{VolSDF}
    \end{subfigure}
    \begin{subfigure}{0.45\linewidth}
        \caption*{Ours}
    \end{subfigure}
    \caption{Rendering and geometric error maps of two DTU scenes~\cite{dtu} (blue=low, yellow=high). Compared to VolSDF~\cite{volsdf} on which we build, our method significantly improves accuracy. 
    }
    \vspace{-8pt}
    \label{fig:dtu_3d}
\end{figure}

\begin{figure*}
    \centering
    \includegraphics[width=0.16\linewidth, trim={100, 0, 160, 0},clip]{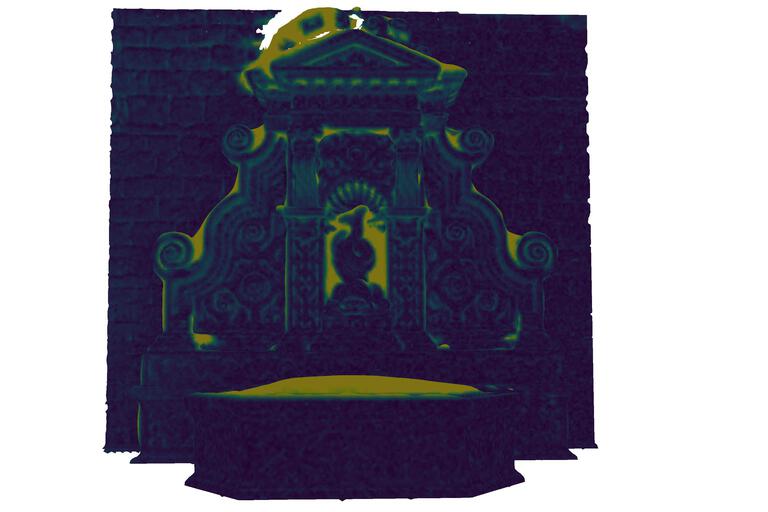}
    \hfill
    \includegraphics[width=0.16\linewidth, trim={100, 0, 160, 0},clip]{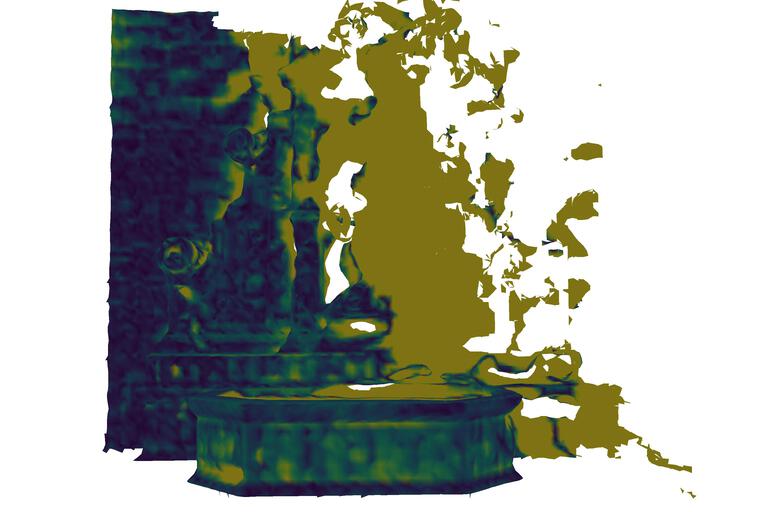}
    \hfill
    \includegraphics[width=0.16\linewidth, trim={100, 0, 160, 0},clip]{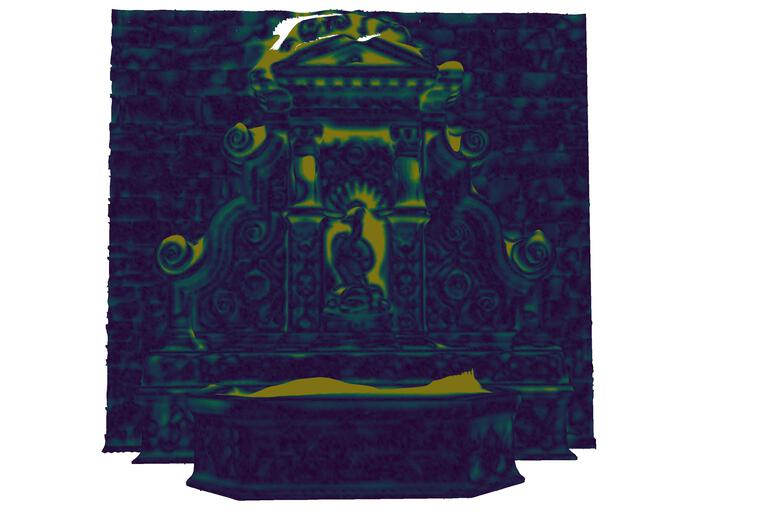}
    \hfill
    \includegraphics[width=0.16\linewidth, trim={100, 0, 160, 0},clip]{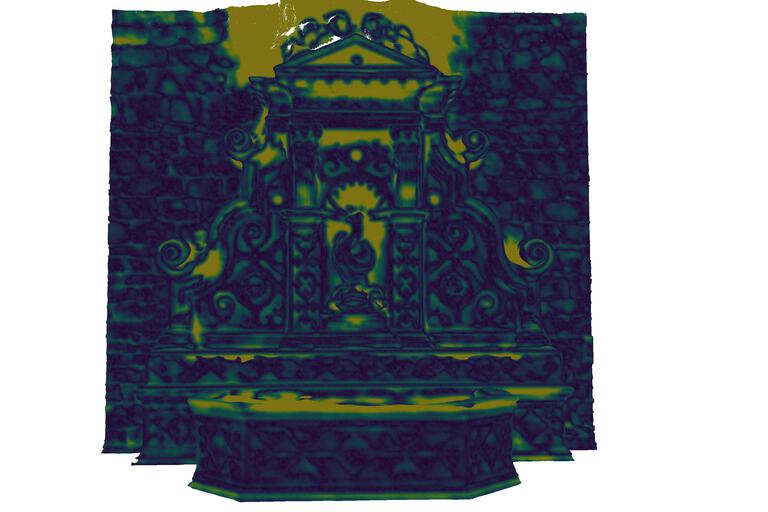}
    \hfill
    \includegraphics[width=0.16\linewidth, trim={100, 0, 160, 0},clip]{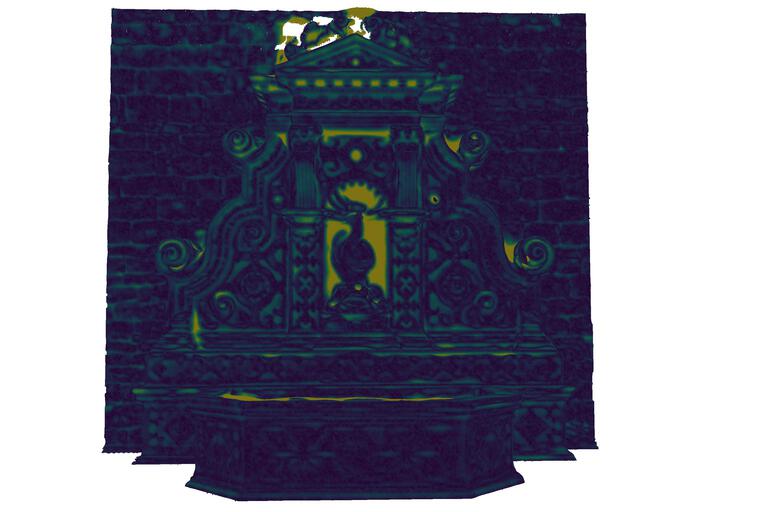} \\
    \includegraphics[width=0.16\linewidth, trim={250, 37, 160, 88},clip]{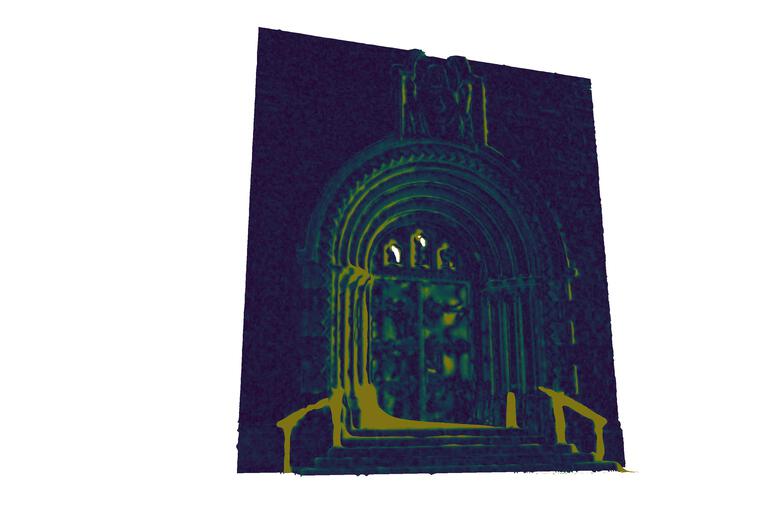}
    \hfill
    \includegraphics[width=0.16\linewidth, trim={250, 37, 160, 88},clip]{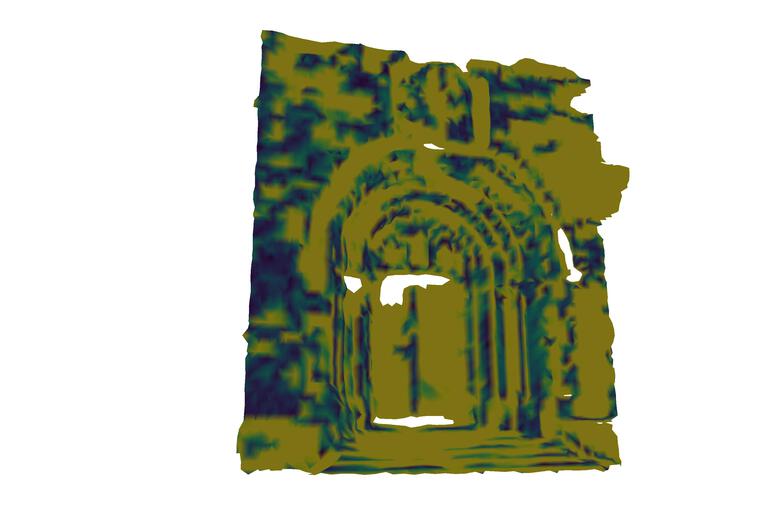}
    \hfill
    \includegraphics[width=0.16\linewidth, trim={250, 37, 160, 88},clip]{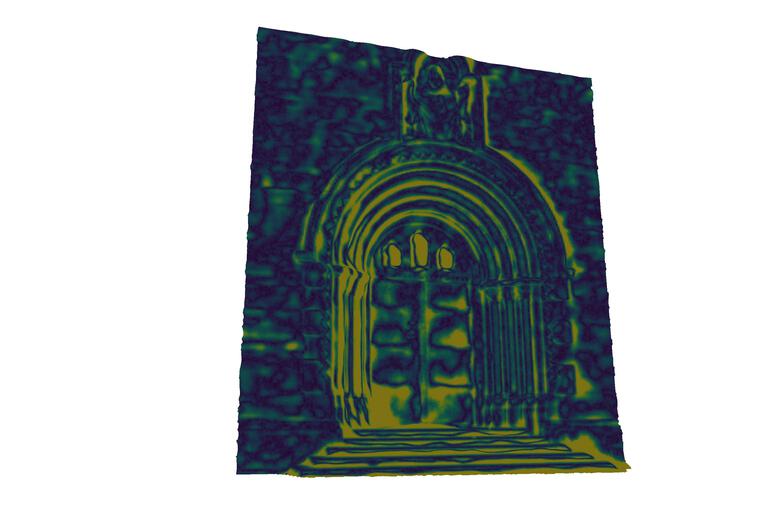}
    \hfill
    \includegraphics[width=0.16\linewidth, trim={250, 37, 160, 88},clip]{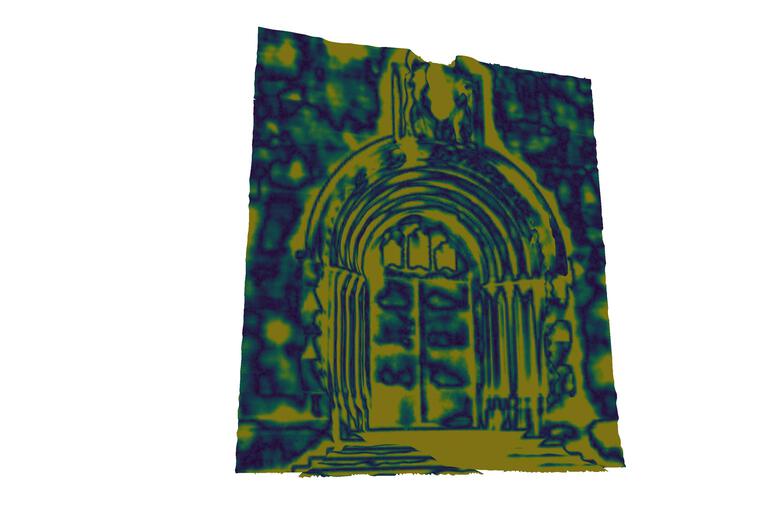}
    \hfill
    \includegraphics[width=0.16\linewidth, trim={250, 37, 160, 88},clip]{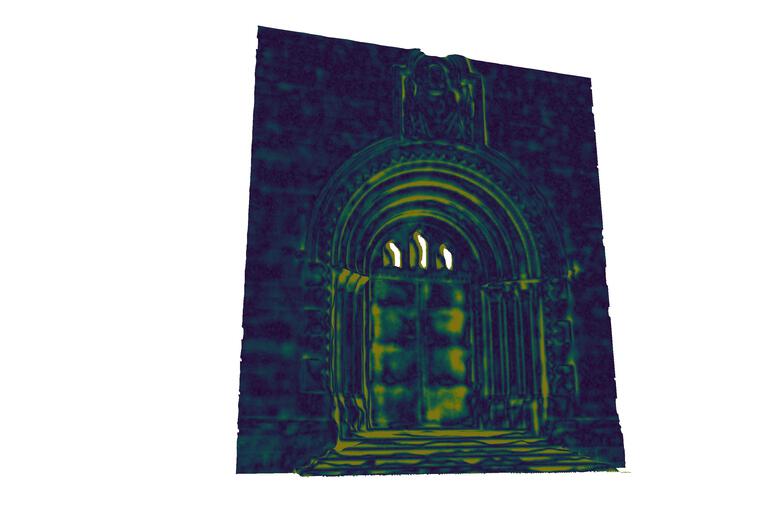}\\
    \begin{subfigure}{0.17\linewidth}
        \caption*{COLMAP~\cite{colmap}}
    \end{subfigure} 
    \hfill
    \begin{subfigure}{0.17\linewidth}
        \caption*{UNISURF~\cite{unisurf}}
    \end{subfigure}
    \hfill
    \begin{subfigure}{0.17\linewidth}
        \caption*{MVSDF~\cite{mvsdf}}
    \end{subfigure}
    \hfill
    \begin{subfigure}{0.17\linewidth}
        \caption*{VolSDF~\cite{volsdf}}
    \end{subfigure}
    \hfill
    \begin{subfigure}{0.17\linewidth}
        \caption*{Ours}
    \end{subfigure}
    \caption{Geometric error maps of reconstructed meshes on EPFL benchmark~\cite{strecha} (blue=low, yellow=high). We ran COLMAP and UNISURF official implementations, MVSDF results were sent by the authors and we ran VolSDF with our implementation. Our method produces better results than other neural implicit surface approaches, almost on par with COLMAP~\cite{colmap}. }
    \vspace{-8pt}
    \label{fig:strecha}
\end{figure*}
    

\section{Experiments}

In this section, we first show that our method outperforms state-of-the-art unsupervised neural implicit surface approaches on the DTU dataset~\cite{dtu}, then on the EPFL benchmark~\cite{strecha}, we present an ablation study to evaluate each of our technical contributions and finally we discuss the limitations of our method. 

\vspace{-8pt}
\paragraph{DTU benchmark:}

The DTU benchmark~\cite{dtu} includes scenes with 49 to 64 images associated to reference point clouds acquired with laser sensor. Each scene covers a different object: some have challenging specular materials while others have large textureless regions. The evaluation of IDR~\cite{idr} selected $15$~scenes and manually annotated object masks. We compare our method with existing work on the same scenes, using DTU evaluation code. The metric is the average of accuracy and completeness: the chamfer distance of prediction to reference point cloud and inversely. Similar to existing work~\cite{idr, unisurf, volsdf, neus}, we clean the output meshes with the visibility masks dilated by 12 pixels.

We compare with multiple baselines in Table~\ref{tab:dtu}. The results for each method are taken from their original paper, except COLMAP, NeRF and IDR that we took from~\cite{unisurf}. Similar to~\cite{unisurf,volsdf,neus}, we only compare in the bottom part of the table deep implicit surface approaches that do not use masks or other data during training. In particular, MVSDF~\cite{mvsdf} uses a supervised depth estimation network. Our method outperforms existing methods by a large margin. As could be expected, the improvement is more important on highly textured scenes but our method performs on par with other methods on weakly textured scenes. \figref{fig:dtu} compares original images, volume rendering obtained by VolSDF, volume rendering obtained with our radiance network and our image warping (using the pixel warping approach described in Section~\ref{sec:warp}). Volumetric rendering only renders smoothed texture, whereas our warping is able to render high-frequency texture information. As can be qualitatively seen in Figures~\ref{fig:teaser} and~\ref{fig:dtu_3d}, this leads to important improvements in accuracy. Reconstructions and geometric error maps for all scenes are shown in the supplementary material.  

\begin{table}[t]
    \small
    \setlength{\tabcolsep}{2pt}
    \centering
    \begin{tabular}{l| c c c c |c c}
    \multirow{2}{*}{Method} & \multicolumn{2}{c}{Fountain-P11} & \multicolumn{2}{c|}{Herzjesu-P7} & \multicolumn{2}{c}{\bf Mean} \\
    & Full & Center & Full & Center & \bf Full & \bf Center\\
    \hline
    COLMAP~\cite{colmap} & \bf 6.47 & \underline{2.45} & \bf 7.95 & \underline{2.31} & \bf 7.21 & \underline{2.38} \\
    UNISURF~\cite{unisurf} & 26.16 & 17.72 & 27.22 & 13.72 & 26.69 & 15.72 \\
    MVSDF~\cite{mvsdf} & \underline{6.87} & 2.26 & 11.32 & 2.72 & 9.10 & 2.49 \\
    VolSDF~\cite{volsdf} & 12.89 & 2.99 & 13.61 & 4.58 & 13.25 & 3.78 \\
    NeuralWarp (ours) & 7.77 & \bf 1.92 & \underline{8.88} & \bf 2.03 & \underline{8.32} & \bf 1.97 \\
    \end{tabular}
    \caption{Quantitative evaluation on EPFL dataset. We use the chamfer distance on the Full scene (Full) and on a manually defined bounding box at the center of the scene (Center). Results of VolSDF~\cite{volsdf} come from our implementation, MVSDF reconstructions were sent by the authors and and we ran MVSDF and COLMAP public implementations. Although COLMAP is the best method for the full reconstruction, our method has the best results for the center metrics. It outperforms existing neural implicit surfaces by 20\% on center metrics.}
    \vspace{-8pt}
    \label{tab:strecha}
\end{table}

\vspace{-8pt}
\paragraph{EPFL benchmark:}

The EPFL benchmark~\cite{strecha} consists of two outdoor scenes of 7 and 11 high resolution images with a high resolution ground truth mesh. 
Since the extent of the ground truth does not exactly overlap the cameras viewing angle and inversely, it is necessary to remove points from both predicted and ground truth meshes for evaluation. 
MVSDF~\cite{mvsdf} uses manual masks to remove vertices from the ground truth mesh, which we argue might be biased. We instead automatically remove vertices from the ground truth mesh when they do not project in any input image. Similar to DTU evaluation, we use silhouette masks to clean the predicted mesh with the scene visual hull. We generate silhouette masks by rendering the ground truth mesh on each input viewpoint and marking pixels which are not covered as outside of the silhouette. Finally, we also remove from the predicted mesh any triangle that is not rendered in any image, which removes in particular faces closing the volume behind the object.  
To compute the distance between the filtered ground truth and predicted mesh, we sample 1 million point from each and compute their chamfer distance.
We call this metrics the \textit{full} chamfer distance. It is mainly influenced by the completeness of the reconstruction, e.g. it compares how well methods reconstruct the ground plane or rarely seen points. 
We therefore introduce another metric referred to as the \textit{center} chamfer distance which only evaluates the chamfer distance in a box at the center of the scene which we manually defined so that it only includes the central part of the scene, which is reconstructed by all methods. Thus, this metric focuses more on the accuracy of the reconstruction. 

We compare our method with several baselines with these two metrics in Table~\ref{tab:strecha}. We ran COLMAP followed by sPSR~\cite{spsr} with trim 5, used the official UNISURF implementation,\footnote{\url{https://github.com/autonomousvision/unisurf}} evaluated the MVSDF meshes communicated by the authors, and ran our own reimplementation of VolSDF. Qualitative comparisons between the reconstructions and error maps for each method can be seen in Figure~\ref{fig:strecha}. Similar to the DTU results, our method outperforms other neural implicit surfaces by more than 20\%.  COLMAP~\cite{colmap} is the best method for the full metric. This is in large part because it is the only method able to reconstruct accurately the ground plane on both scenes. For the center metrics however, our method outperforms even COLMAP, though this might be due to some details reconstructed by COLMAP (e.g. railing) not being included in the ground truth: qualitatively, COLMAP still seems to recover finer details. 

\begin{table}[t]
    \small
    \centering
    \begin{tabularx}{\linewidth}{X |c c c c}
        Method & $\mathcal{L}_\text{vol}$ & $\mathcal{L}_\text{warp}$ & $M_s^{\text{occ}}$ & Chamfer dist. \\
        \hline
        VolSDF~\cite{volsdf} & \checkmark & None &  & 0.85\\
        Pixel & \checkmark & Pixel & \checkmark & 0.83\\
        Patch no occ. & \checkmark & Patch &  & 0.74\\
        Patch no vol. & & Patch & \checkmark & 0.74\\
        NeuralWarp (full) & \checkmark & Patch & \checkmark & \bf 0.68\\
    \end{tabularx}
    \caption{Ablation study on all scenes of DTU: $\mathcal{L}_\text{vol}$ denotes volumetric rendering, $\mathcal{L}_\text{warp}$ warping consistency for which we try none, pixel and patch warpings. $M_s^{\text{occ}}$ denotes whether we detect self occlusions or not.}
    \vspace{-8pt}
    \label{tab:ablation}
\end{table}

\vspace{-8pt}
\paragraph{Ablation study:}

To evaluate the effect of our technical contributions, we perform an ablation study on the DTU dataset. Starting from the same models trained without photometric consistency, we finetune different versions of our model for 50000 iterations and compare the results.  The average chamfer distance over all 15 scenes is shown in Table~\ref{tab:ablation} and we report the results on each scene in the supplementary material.  We first compare the results without our warping loss ('VolSDF~\cite{volsdf}' line), with pixel warping ('Pixel' line) and with patch warping ('NeuralWarp (full)' line). Both pixel and patch warping improve the results, with a clear advantage for patches. We then evaluate the importance of masking. Removing the projection mask (not reported in the table) does not lead to meaningful reconstructions. Without the occlusion mask ('Patch no occ.' line) our method still improves over the baseline, but by a smaller margin. Finally, we tried to completely remove volumetric rendering loss, that is, use $\lambda_{\text{vol}} = 0$ ('Patch no vol.' line). Again, this improves over the baseline but is worse than combining the volumetric and warp losses.

\vspace{-8pt}
\paragraph{Limitations:} Our method has several limitations. First, compared to COLMAP, it struggles to reconstruct high-frequency geometry. We believe this is due to the difficulty of optimizing a geometry network at high resolution. Second, computing our loss increases the computational cost of the optimization, in particular the occlusion mask adds processing time and processing patches increases memory footprint. Finally, simply comparing patches does not model reflections, which can lead to artifacts even using a robust patch similarity. We show such an example in supplementary material.

\section{Conclusion}

We have presented a new method to perform multiview reconstruction with implicit functions, using image warpings in combination with volumetric rendering. Unlike existing neural implicit surface methods, our approach can easily take advantage of high-frequency texture. We show this leads to strong performance improvements on the classical DTU and EPFL datasets. 

\section*{Acknowledgments}

\small
This work was supported in part by ANR project EnHerit ANR-17-CE23-0008 and was performed using HPC resources from GENCI–IDRIS 2021-AD011011756R1. We thank Tom Monnier and Bruno Lecouat for valuable feedback and Jingyang Zhang for sending MVSDF results.

{\small
\bibliographystyle{ieee_fullname}
\bibliography{egbib}
}

\end{document}


\title{Supplementary material \\Improving neural implicit surfaces geometry with patch warping}

\maketitle

This document first show a failure case of our method. It then presents the detailed results of the ablation study and additional qualitative results on DTU dataset.

\section{Failure case}

\begin{figure}
    \centering
        \begin{subfigure}{0.01\textwidth}
        \rotatebox{90}{\quad \quad \quad \quad VolSDF \quad} \\
        \rotatebox{90}{\quad \quad \quad \quad \quad Ours \quad \quad \quad \quad \quad} \\
    \end{subfigure}
    \begin{subfigure}{0.3\textwidth}
        \includegraphics[width=\linewidth]{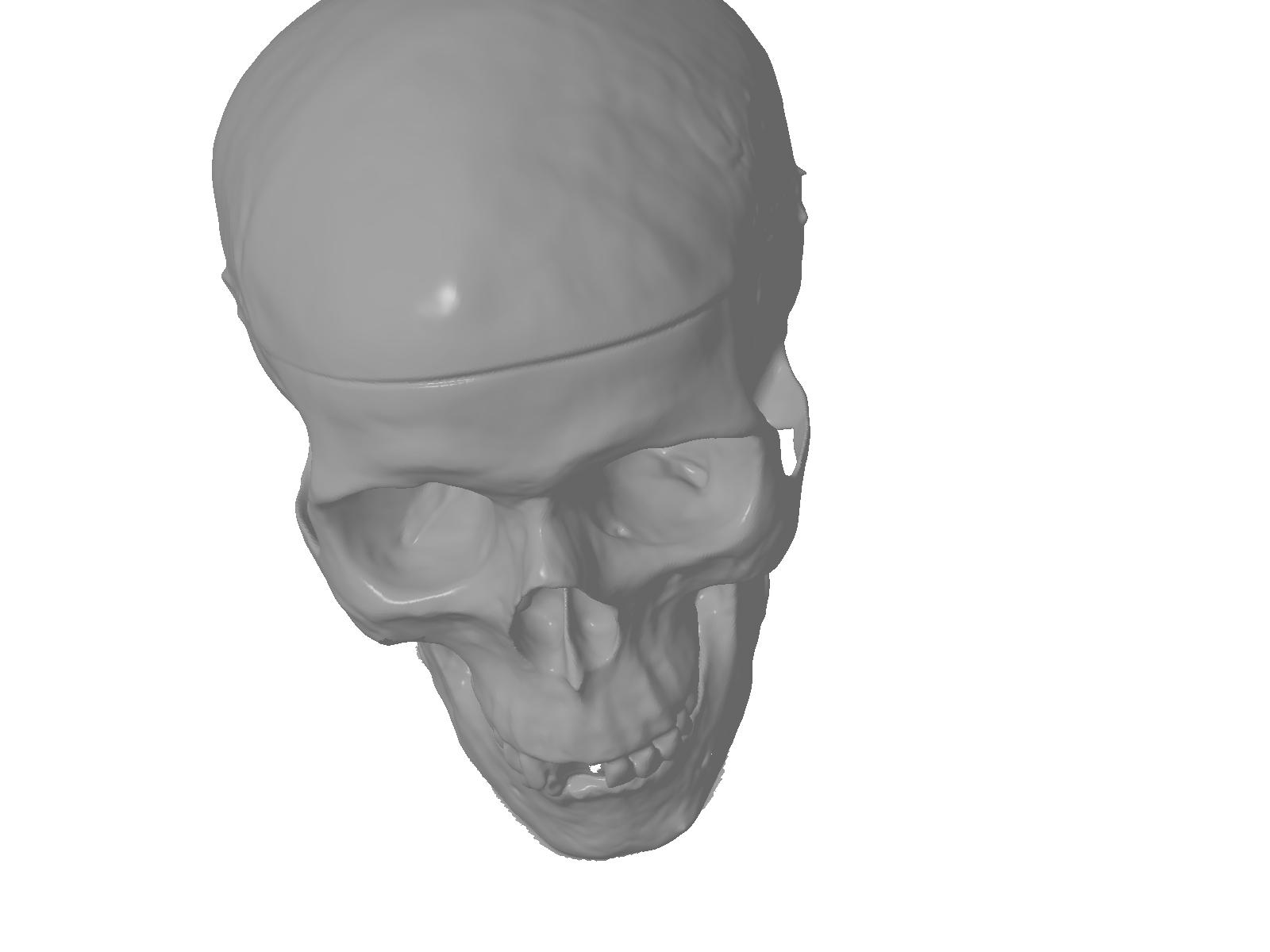}
        \includegraphics[width=\linewidth]{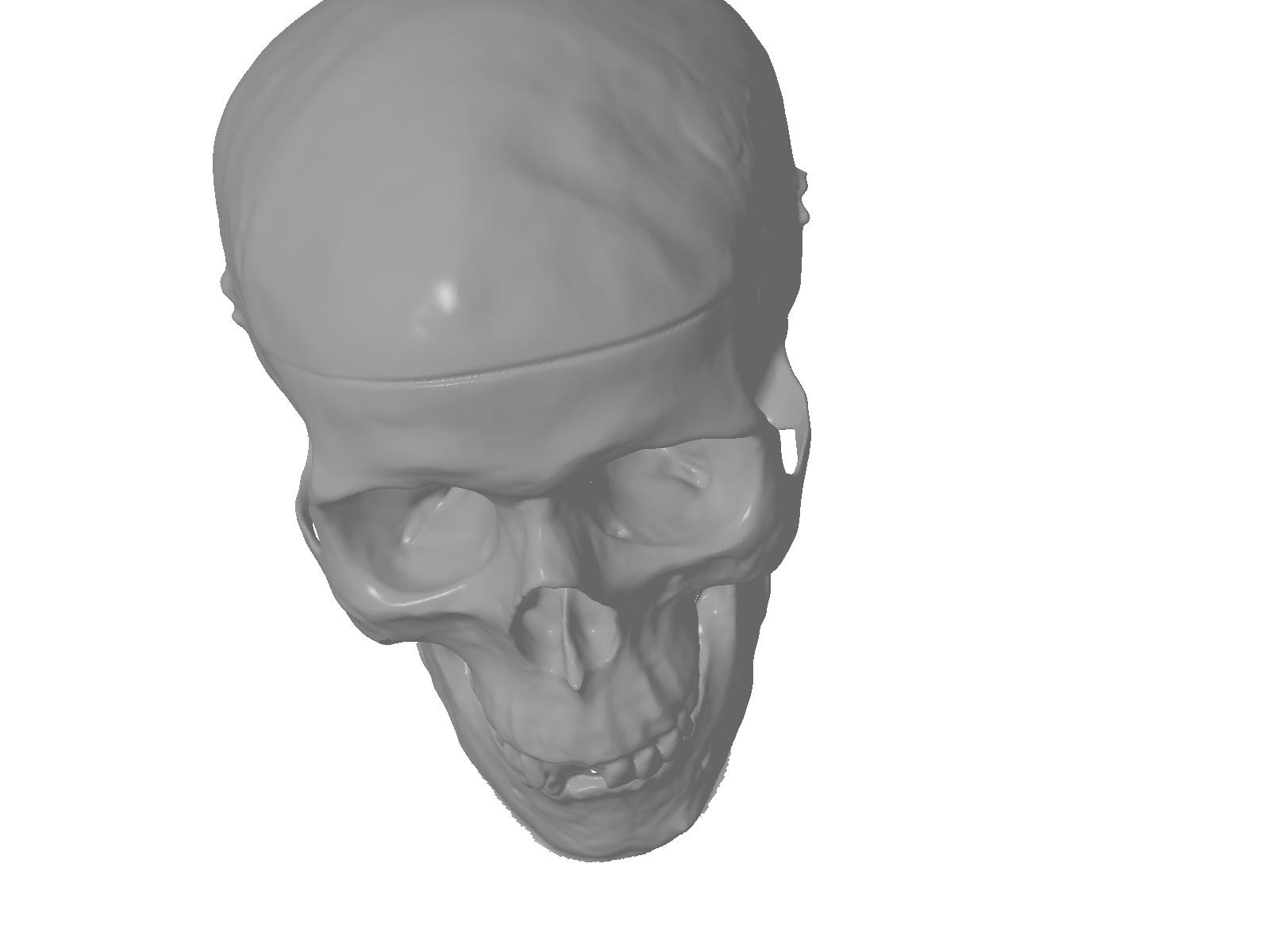}
        \caption{Reconstructions}
    \end{subfigure}
    \hfill
    \begin{subfigure}{0.3\textwidth}
        \includegraphics[width=\linewidth]{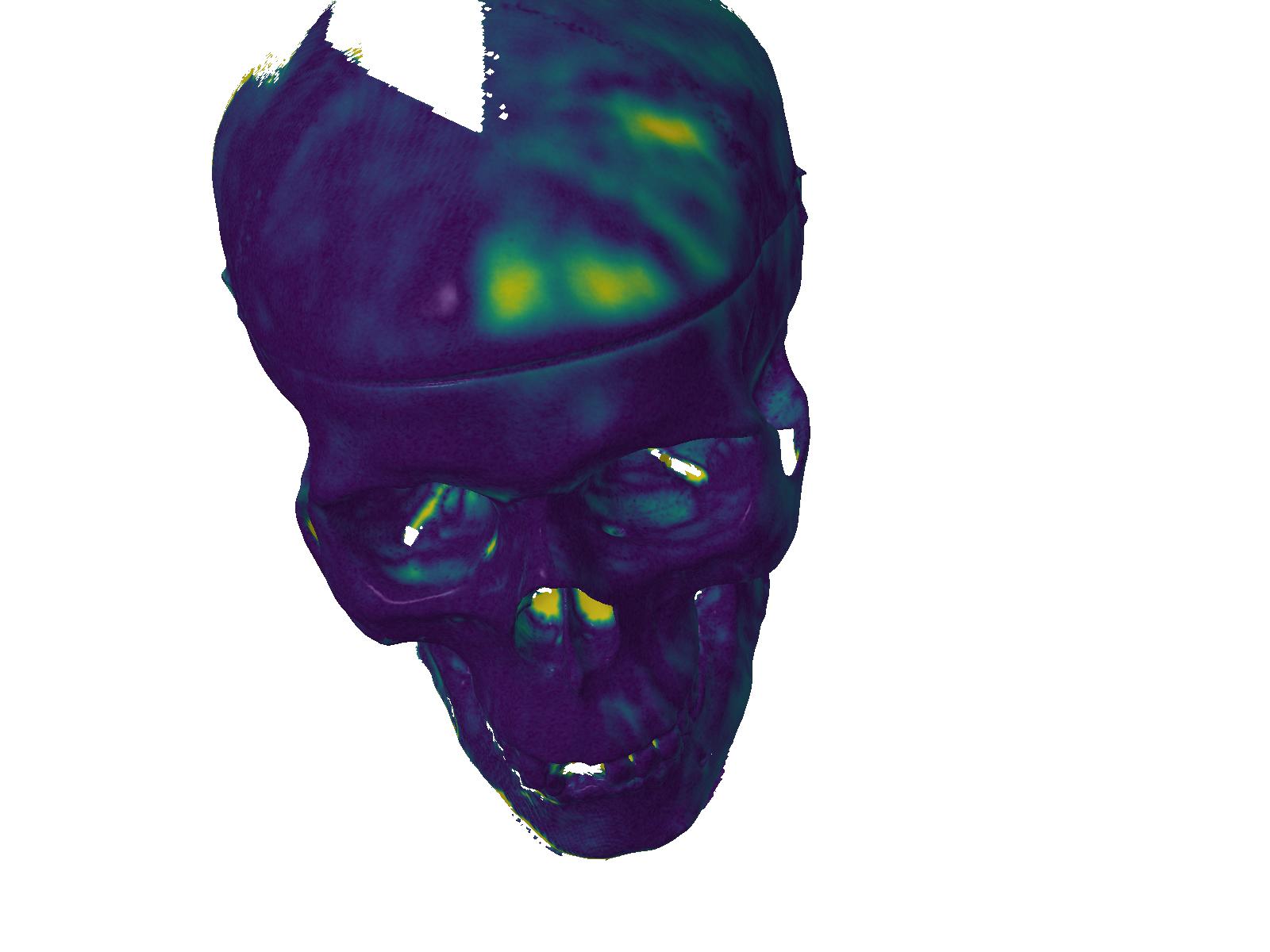}
        \includegraphics[width=\linewidth]{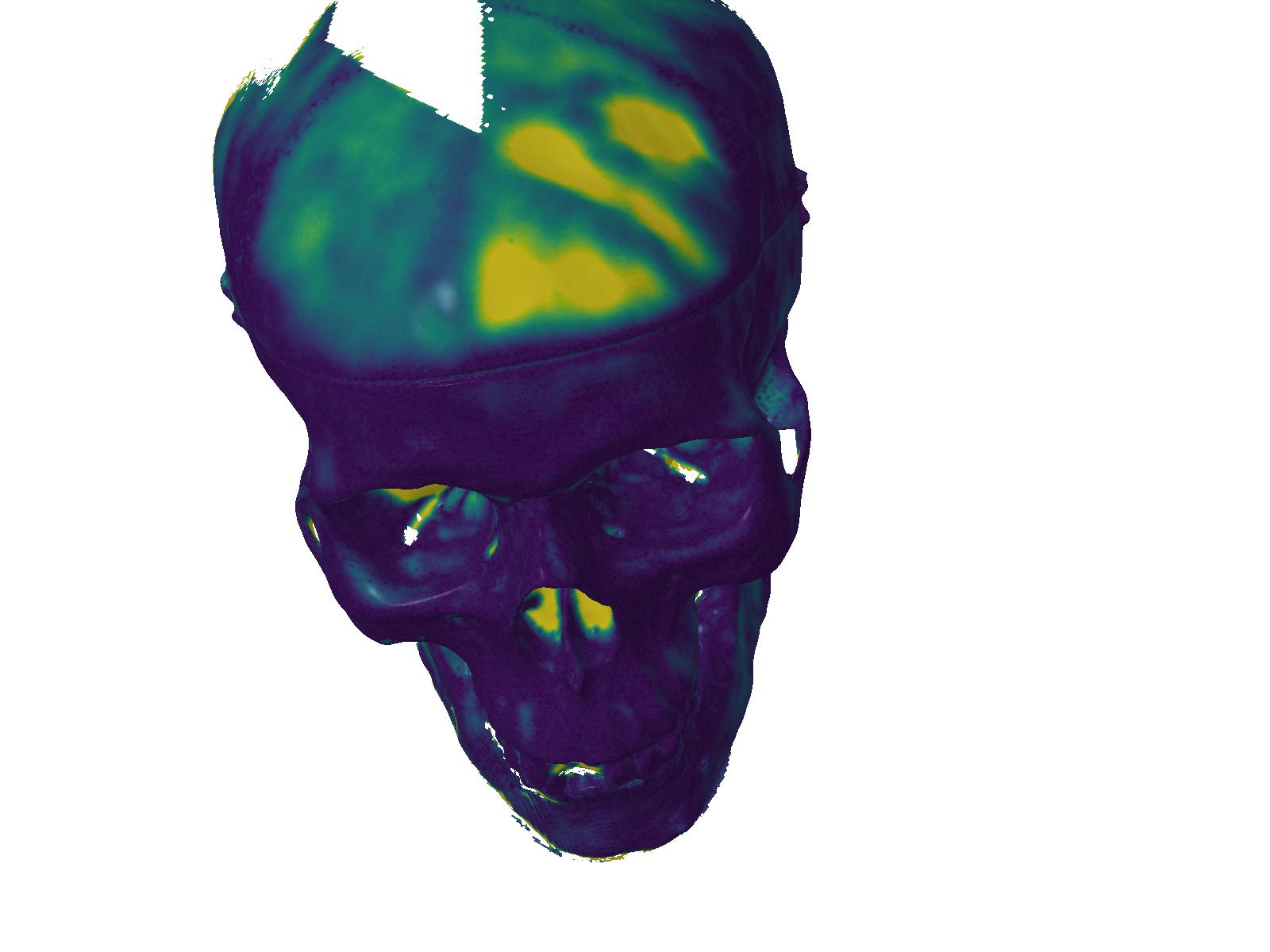}
        \caption{Accuracy colormaps}
    \end{subfigure}
    \hfill
    \begin{subfigure}{0.3\textwidth}
        \includegraphics[width=\linewidth, trim=0 -21.2cm 0 -21.2cm, clip]{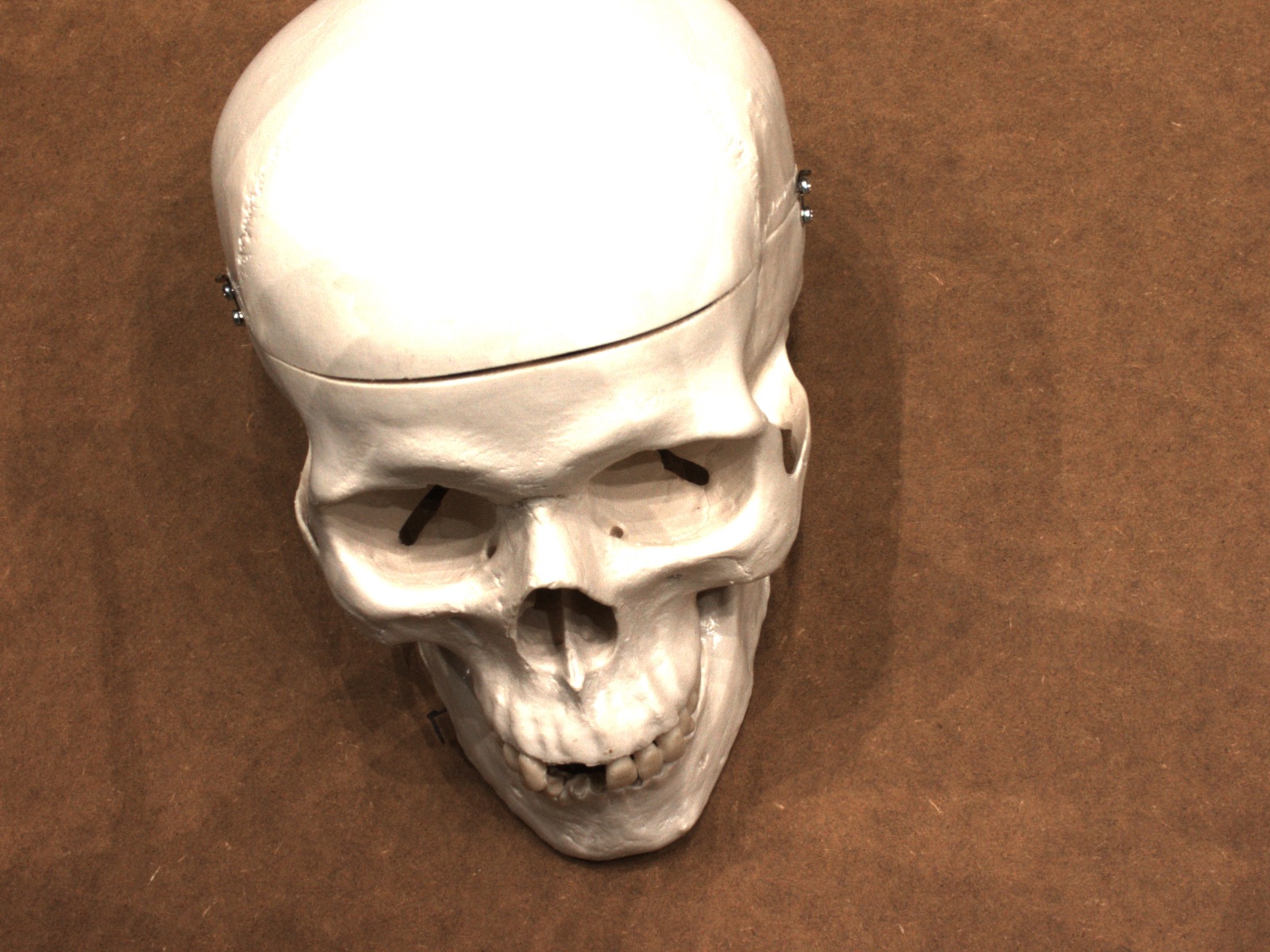}
        \caption{Input image}
    \end{subfigure}
    \caption{Failure case for a scene with low texture and large view dependent effects.}
    \label{fig:fail_case}
\end{figure}

Figure~\ref{fig:fail_case} shows reconstructions from scan 65 of DTU which is a failure case of our method. This scene has large view dependent effects like specularities or saturated pixels. Optimizing consistency from multiple view produces artifacts in the reconstruction since there is no useful signal. This may explain why reconstructions optimized with volumetric rendering only are smoother and more accurate than the reconstruction with our full method. 

\section{Detailed results ablation study}

We show in tables~\ref{tab:ablation_supp_a}, ~\ref{tab:ablation_supp_b} and ~\ref{tab:ablation_supp_c}  the three quality metrics (overall, accuracy and completeness) for each scene of the ablation study. Adding volumetric rendering is especially beneficial on scenes with poor texture and large specularities like scans 63, 65, 97 and 110.

\begin{table}[h]
    \centering
    \small
    \setlength{\tabcolsep}{3pt}
    \begin{tabular}{c c c c | c c c c c c c c c c c c c c c| c}
    Method & $\mathcal{L}_\text{vol}$ & $\mathcal{L}_\text{warp}$ & $M_s^{\text{occ}}$ & 24 & 37 & 40 & 55 & 63 & 65 & 69 & 83 & 97 & 105 & 106 & 110 & 114 & 118 & 122 & Mean \\
    \hline
    VolSDF & \checkmark &  &  & 0.99 & 0.86 & 0.82 & 0.48 & 1.39 & \bf 0.69 & 0.90 & 1.26 & 1.24 & 0.67 & 0.67 & 1.12 & 0.41 & 0.72 & 0.54 & 0.85\\
    Pixel & \checkmark & Pixel & \checkmark & 0.66 & 0.77 & 0.65 & 0.46 & 1.36 & 0.85 & 0.83 & 1.26 & 1.15 & 0.69 & 0.65 & 1.46 & 0.46 & 0.66 & 0.55 & 0.83\\
    Patch no occ. & \checkmark & Patch &  & 0.58 & \bf 0.69 & 0.41 & 0.42 & 0.84 & 0.92 & 0.87 & 1.30 & 1.24 & 0.70 & 0.69 & 0.81 & \bf 0.38 & 0.69 & 0.52 & 0.74\\
    Patch no vol. &  & Patch & \checkmark & 0.49 & 0.78 & 0.41 & \bf 0.33 & 0.92 & 1.04 & \bf 0.82 & 1.25 & 1.29 & \bf 0.67 & \bf 0.62 & 0.93 & 0.40 & 0.67 & \bf 0.50 & 0.74\\
    Ours full & \checkmark & Patch & \checkmark & \bf 0.49 & 0.71 & \bf 0.38 & 0.38 & \bf 0.79 & 0.81 & 0.82 & \bf 1.20 & \bf 1.06 & 0.68 & 0.66 & \bf 0.74 & 0.41 & \bf 0.63 & 0.51 & \bf 0.68\\
    \end{tabular}
    \vspace{-5pt}
    \caption{Detailed \textbf{overall} metric of ablation study on DTU}
    \vspace{10pt}
    \label{tab:ablation_supp_a}

    \begin{tabular}{c c c c | c c c c c c c c c c c c c c c| c}
    Method & $\mathcal{L}_\text{vol}$ & $\mathcal{L}_\text{warp}$ & $M_s^{\text{occ}}$ & 24 & 37 & 40 & 55 & 63 & 65 & 69 & 83 & 97 & 105 & 106 & 110 & 114 & 118 & 122 & Mean \\
    \hline
    VolSDF & \checkmark &  &  & 0.90 & 0.92 & 0.77 & 0.51 & 1.31 & \bf 0.70 & 0.96 & 0.88 & 1.00 & 0.62 & 0.61 & 1.24 & 0.39 & 0.77 & 0.60 & 0.81\\
    Pixel & \checkmark & Pixel & \checkmark & 0.65 & 0.86 & 0.69 & 0.49 & 1.37 & 0.84 & \bf 0.89 & 0.89 & 0.91 & 0.61 & \bf 0.54 & 1.56 & 0.44 & \bf 0.70 & 0.59 & 0.80\\
    Patch no occ. & \checkmark & Patch &  & 0.65 & \bf 0.79 & 0.41 & 0.51 & 1.06 & 0.91 & 0.97 & 0.92 & 0.99 & 0.63 & 0.61 & 1.04 & \bf 0.37 & 0.73 & \bf 0.54 & 0.74\\
    Patch no vol. &  & Patch & \checkmark & 0.54 & 0.93 & 0.43 & \bf 0.36 & 1.24 & 1.07 & 0.92 & 0.91 & 1.05 & 0.61 & 0.56 & 1.21 & 0.42 & 0.80 & 0.57 & 0.77\\
Ours full & \checkmark & Patch & \checkmark & \bf 0.52 & 0.82 & \bf 0.39 & 0.40 & \bf 1.01 & 0.81 & 0.92 & \bf 0.85 & \bf 0.84 & \bf 0.59 & 0.57 & \bf 0.92 & 0.41 & 0.73 & 0.55 & \bf 0.69\\
    \end{tabular}
    \vspace{-5pt}
    \caption{Detailed \textbf{accuracy} metric of ablation study on DTU}
    \vspace{10pt}
    \label{tab:ablation_supp_b}

    \begin{tabular}{c c c c | c c c c c c c c c c c c c c c| c}
    Method & $\mathcal{L}_\text{vol}$ & $\mathcal{L}_\text{warp}$ & $M_s^{\text{occ}}$ & 24 & 37 & 40 & 55 & 63 & 65 & 69 & 83 & 97 & 105 & 106 & 110 & 114 & 118 & 122 & Mean \\
    \hline
    VolSDF & \checkmark &  &  & 1.09 & 0.79 & 0.86 & 0.44 & 1.46 & \bf 0.68 & 0.84 & 1.64 & 1.49 & \bf 0.73 & 0.73 & 1.00 & 0.43 & 0.66 & 0.48 & 0.89\\
    Pixel & \checkmark & Pixel & \checkmark & 0.68 & 0.69 & 0.62 & 0.44 & 1.36 & 0.86 & 0.77 & 1.63 & 1.39 & 0.77 & 0.76 & 1.37 & 0.48 & 0.61 & 0.50 & 0.86\\
    Patch no occ. & \checkmark & Patch &  & 0.52 & \bf 0.60 & 0.40 & 0.32 & 0.63 & 0.92 & 0.76 & 1.68 & 1.48 & 0.77 & 0.77 & 0.59 & \bf 0.39 & 0.66 & 0.50 & 0.73\\
    Patch no vol. &  & Patch & \checkmark & \bf 0.45 & 0.64 & 0.39 & \bf 0.31 & 0.60 & 1.01 & \bf 0.71 & 1.59 & 1.52 & 0.73 & \bf 0.69 & 0.66 & 0.39 & \bf 0.54 & \bf 0.43 & 0.71\\
    Ours full & \checkmark & Patch & \checkmark & 0.46 & 0.61 & \bf 0.37 & 0.37 & \bf 0.58 & 0.81 & 0.72 & \bf 1.55 & \bf 1.28 & 0.77 & 0.74 & \bf 0.56 & 0.40 & 0.54 & 0.46 & \bf 0.68\\
    \end{tabular}
    \vspace{-5pt}
    \caption{Detailed \textbf{completeness} metric of ablation study on DTU}
    \label{tab:ablation_supp_c}
\end{table}

\newpage
\section{Additional DTU visualizations}

We show additional DTU results in figures~\ref{fig:supp_vis_a}, ~\ref{fig:supp_vis_b} and ~\ref{fig:supp_vis_c}

\begin{figure}[h]
    \centering
    \includegraphics[width=\textwidth]{suppfigs/suppfig1.pdf}
    \begin{subfigure}[b]{0.24\textwidth}
        \caption*{VolSDF reconstruction}
    \end{subfigure}
    \hfill
    \begin{subfigure}[b]{0.24\textwidth}
        \caption*{Ours reconstruction}
    \end{subfigure}
    \hfill
    \begin{subfigure}[b]{0.24\textwidth}
        \caption*{VolSDF error map}
    \end{subfigure}
    \hfill
    \begin{subfigure}[b]{0.24\textwidth}
        \caption*{Ours error map}
    \end{subfigure}
    \caption{Qualitative comparison of VolSDF and our method (1/3)}
    \label{fig:supp_vis_a}
\end{figure}

\begin{figure}[h]
    \centering
    \includegraphics[width=\textwidth]{suppfigs/suppfig2.pdf}
    \begin{subfigure}[b]{0.24\textwidth}
        \caption*{VolSDF reconstruction}
    \end{subfigure}
    \hfill
    \begin{subfigure}[b]{0.24\textwidth}
        \caption*{Ours reconstruction}
    \end{subfigure}
    \hfill
    \begin{subfigure}[b]{0.24\textwidth}
        \caption*{VolSDF error map}
    \end{subfigure}
    \hfill
    \begin{subfigure}[b]{0.24\textwidth}
        \caption*{Ours error map}
    \end{subfigure}
    \caption{Qualitative comparison of VolSDF and our method (2/3)}
    \label{fig:supp_vis_b}
\end{figure}

\begin{figure}[h]
    \centering
    \includegraphics[width=\textwidth]{suppfigs/suppfig3.pdf}
    \begin{subfigure}[b]{0.24\textwidth}
        \caption*{VolSDF reconstruction}
    \end{subfigure}
    \hfill
    \begin{subfigure}[b]{0.24\textwidth}
        \caption*{Ours reconstruction}
    \end{subfigure}
    \hfill
    \begin{subfigure}[b]{0.24\textwidth}
        \caption*{VolSDF error map}
    \end{subfigure}
    \hfill
    \begin{subfigure}[b]{0.24\textwidth}
        \caption*{Ours error map}
    \end{subfigure}
    \caption{Qualitative comparison of VolSDF and our method (3/3)}
    \label{fig:supp_vis_c}
\end{figure}